\documentclass[journal]{IEEEtran}
\usepackage{amsmath}
\usepackage{graphicx}
\usepackage{subfigure}
\usepackage{caption2}
\usepackage{longtable}
\usepackage{booktabs}
\usepackage{multirow}
\usepackage{color}
\usepackage{float}
\usepackage[switch]{lineno}

\ifCLASSINFOpdf
\else
\fi
\begin{document}
%
\title{Sharp Eyes: A Salient Object Detector Working The Same Way as Human Visual Characteristics}

%
%
%

\author{Ge Zhu,
        Jinbao Li,~\IEEEmembership{Member,~IEEE,}
        and Yahong Guo
\thanks{Ge Zhu is with the School of Electronic Engineering, and the School of Data Science and Technology, Heilongjiang University, Harbin, Heilongjiang, 150080, China (e-mail: zhuge@hlju.edu.cn).}
\thanks{Jinbao Li is with the Shandong Artificial Intelligence Institute, School
of Mathematics and Statistics, Qilu University of Technology (Shandong Academy of Sciences), Jinan, Shandong, 250353, China (e-mail: lijinb@sdas.org). \textit{(Corresponding author: Jinbao Li.)}}
\thanks{Yahong Guo is with the School of Computer Science and Technology, Qilu University of Technology (Shandong Academy of Sciences), Jinan, Shandong, 250353, China (e-mail: guoyh@qlu.edu.cn). \textit{(Corresponding author: Yahong Guo.)}}
}

%
%

\markboth{Journal of \LaTeX\ Class Files,~Vol.~14, No.~8, August~2015}%
{Shell \MakeLowercase{\textit{et al.}}: Bare Demo of IEEEtran.cls for IEEE Journals}
%



\maketitle

\begin{abstract}
Current methods aggregate multi-level features or introduce edge and skeleton to get more refined saliency maps. However, little attention is paid to how to obtain the complete salient object in cluttered background, where the targets are usually similar in color and texture to the background. To handle this complex scene, we propose a sharp eyes network (SENet) that first seperates the object from scene, and then finely segments it, which is in line with human visual characteristics, i.e., to look first and then focus. Different from previous methods which directly integrate edge or skeleton to supplement the defects of objects, the proposed method aims to utilize the expanded objects to guide the network obtain complete prediction. Specifically, SENet mainly consists of target separation (TS) brach and object segmentation (OS) branch trained by minimizing a new hierarchical difference aware (HDA) loss. In the TS branch, we construct a fractal structure to produce saliency features with expanded boundary via the supervision of expanded ground truth, which can enlarge the detail difference between foreground and background. In the OS branch, we first aggregate multi-level features to adaptively select complementary components, and then feed the saliency features with expanded boundary into aggregated features to guide the network obtain complete prediction. Moreover, we propose the HDA loss to further improve the structural integrity and local details of the salient objects, which assigns weight to each pixel according to its distance from the boundary hierarchically. Hard pixels with similar appearance in border region will be given more attention hierarchically to emphasize their importance in completeness prediction. Comprehensive experimental results on five benchmark datasets demonstrate that the proposed approach outperforms the state-of-the-art methods both quantitatively and qualitatively. Code has been made available at https://github.com/lesonly/SENet.
\end{abstract}

\begin{IEEEkeywords}
Computer vision, salient object detection, deep learning, convolutional neural network.
\end{IEEEkeywords}

%
\IEEEpeerreviewmaketitle

\section{Introduction}
%
%
%
%
\IEEEPARstart{S}{alient} object detection (SOD) is commonly interpreted in computer vision as a process that includes two stages:1) detecting the most salient object and 2) segmenting the accurate region of that object \cite{DBLP:journals/cvm/BorjiCHJL19}. As a pre-processing procedure, SOD is widely used in downstream vision tasks, such as image understanding \cite{DBLP:journals/tgrs/ZhangDZ15a}, object tracking \cite{DBLP:conf/icml/HongYKH15} and image retrieval \cite{DBLP:journals/tmm/GaoSTX15}. 
\par In recent years, with the breakthrough of deep learning technology, many methods [5], [6], [7], [8], [9] based on convolution neural network (CNN) have greatly boosted the performance of salient object detection. However, in the process of extracting image features, CNN processes shape, color, texture and brightness together. In some complex scenes with cluttered background which are often similar in color or texture to the foreground, the predicted saliency maps still suffer from incomplete predictions. As shown in Fig. 1, incomplete prediction means that only a part of the salient object can be predicted (rows 1), or the object cannot be highlighted uniformly (row 3, column 5), or no object can be detected (row 2, column 3).
\begin{figure}
	\makeatletter
	\renewcommand{\@thesubfigure}{\hskip\subfiglabelskip}
	\makeatother
	\subfigure[Image]{
		\begin{minipage}[b]{0.135\linewidth}	
			\includegraphics[width=1.16\linewidth]{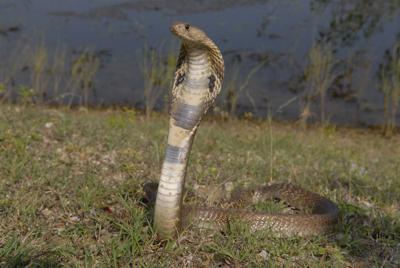}\vspace{1.5pt}
			\includegraphics[width=1.16\linewidth]{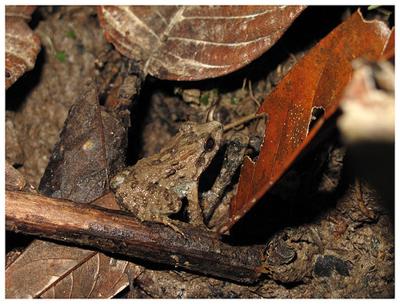}\vspace{1.5pt}
			\includegraphics[width=1.16\linewidth]{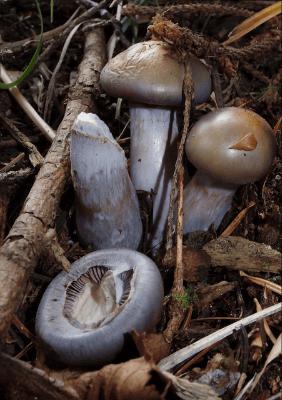}	
	\end{minipage}}
	\subfigure[GT]{
		\begin{minipage}[b]{0.135\linewidth}
			\includegraphics[width=1.16\linewidth]{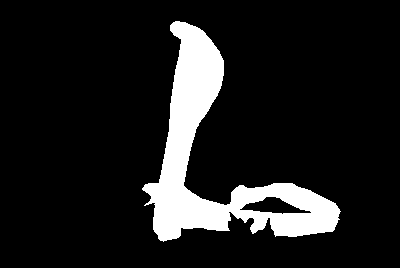}\vspace{1.5pt}			
			\includegraphics[width=1.16\linewidth]{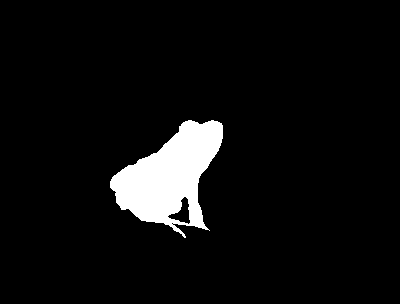}\vspace{1.5pt}
			\includegraphics[width=1.16\linewidth]{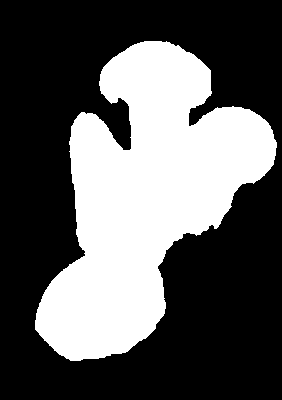}
	\end{minipage}}
	\subfigure[PurNet]{
		\begin{minipage}[b]{0.135\linewidth}	
			\includegraphics[width=1.16\linewidth]{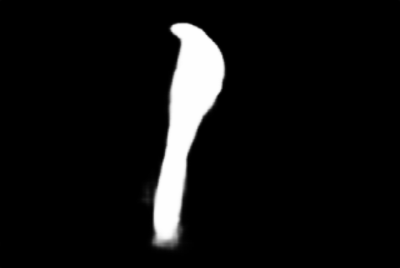}\vspace{1.5pt}	
			\includegraphics[width=1.16\linewidth]{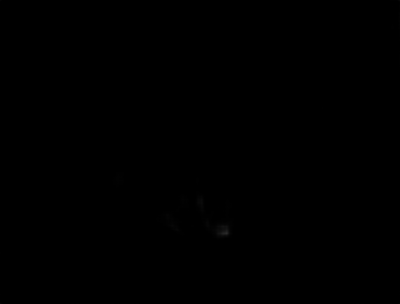}\vspace{1.5pt}
			\includegraphics[width=1.16\linewidth]{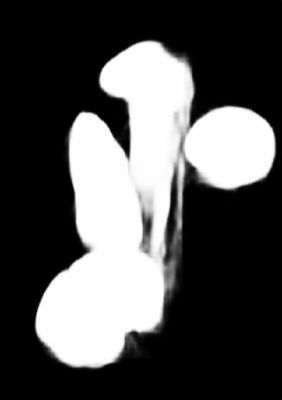}
	\end{minipage}}
	\subfigure[DCN]{
		\begin{minipage}[b]{0.135\linewidth}	
			\includegraphics[width=1.16\linewidth]{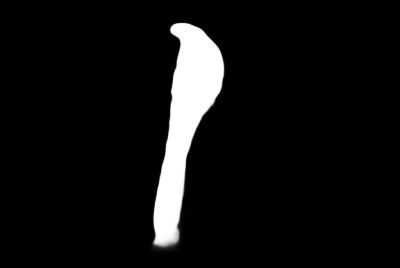}\vspace{1.5pt}		
			\includegraphics[width=1.16\linewidth]{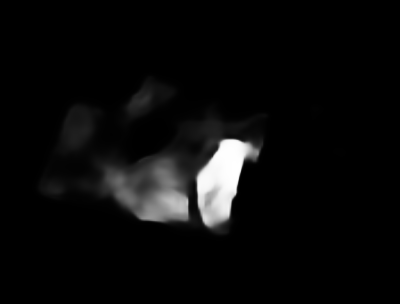}\vspace{1.5pt}
			\includegraphics[width=1.16\linewidth]{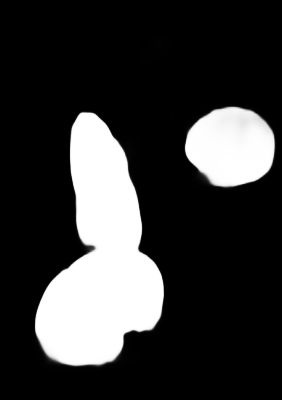}
	\end{minipage}}
	\subfigure[ICON]{
		\begin{minipage}[b]{0.135\linewidth}	
			\includegraphics[width=1.16\linewidth]{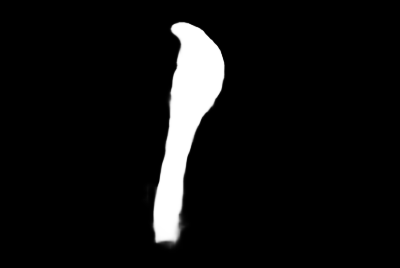}\vspace{1.5pt}	
			\includegraphics[width=1.16\linewidth]{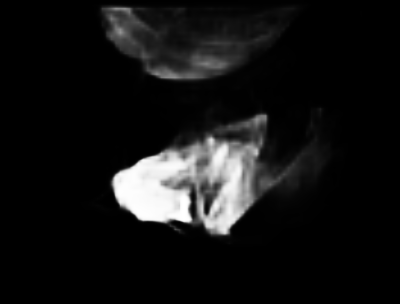}\vspace{1.5pt}
			\includegraphics[width=1.16\linewidth]{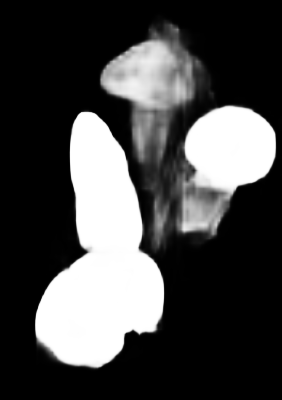}			
	\end{minipage}}
	\subfigure[Ours]{
		\begin{minipage}[b]{0.135\linewidth}	
			\includegraphics[width=1.16\linewidth]{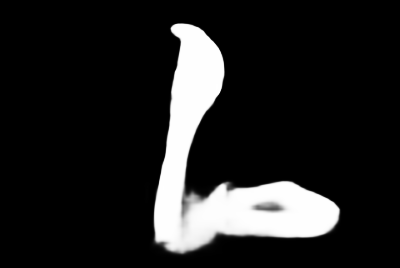}\vspace{1.5pt}		
			\includegraphics[width=1.16\linewidth]{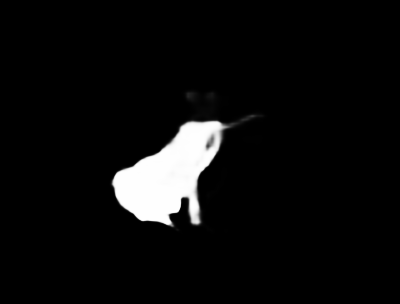}\vspace{1.5pt}
			\includegraphics[width=1.16\linewidth]{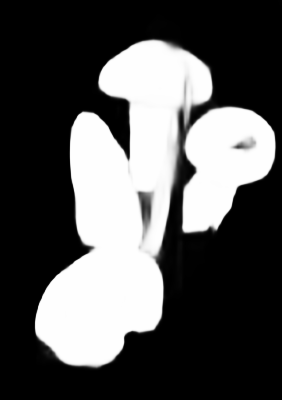}			
	\end{minipage}}
	\caption{Sample results predicted by different methods in complex scenes with cluttered background. In the first row, the snake's body is embedded in the weeds with a similar appearance to it. In the second row, the frog perches on leaves that are similar in color and texture to it. In the third row, the mushrooms are scattered in the thickets with low contrast. However, three recent methods (PurNet \cite{DBLP:journals/tip/LiSXMT21}, DCN \cite{DBLP:journals/tip/WuSH21} and ICON \cite{zhuge2021salient}) obtain incomplete predictions in above scenes, while our method can cope with these complex scenes and generate more complete saliency maps.}
\end{figure}
\par  To get the fine saliency maps, current state-of-the-arts methods mainly focus on how to integrate multi-level features or introduce auxiliary information (i.e., edge and skeleton) to obtain complementary features. To learn comprehensive features, Zhang et al. \cite{DBLP:conf/mm/ZhangLPYL21} and Zhuge et al. \cite{zhuge2021salient} proposed the aggregation networks to integrate multi-level features. However, due to the repeated sub-sampling operation in CNN model, the detail of the salient objects with complex structures will be gradually weaken, which can hardly uniformly highlight the entire salient objects. To obtain the precise boundaries, Zhou et al. \cite{DBLP:conf/cvpr/ZhouXLCY20} and Wei et al. \cite{DBLP:conf/cvpr/WeiWWSH020} introduced the edge information into models, but the edge information only contains a small part of the entire object, which is difficult to fill the interior of objects. To obtain more complementary features, Liu et al. \cite{DBLP:journals/tip/LiuHC20} and Wu et al. \cite{DBLP:journals/tip/WuSH21} integrated both edge and skeleton information into networks. However, the divergence of feature domains may bring background noises, which is difficult to assistant network generate complete saliency maps. Since 100\% recall can be achieved by setting the whole saliency map to be foreground, the challenge is how to solve the problem of incomplete prediction under the premise of low noises in a complex scene. In brief, the above three kinds of methods are all from the perspective of supplementary information. When encountering the cluttered background, it is difficult to separate the target completely. Fig. 1 shows the sample results of different methods in cluttered scenes, including the PurNet \cite{DBLP:journals/tip/LiSXMT21} based on using purificatory mechanism, DCN \cite{DBLP:journals/tip/WuSH21} based on introducing the boundary and skeleton information, and ICON \cite{zhuge2021salient} based on aggregating multi-level features. 
\par In addition, most of existing models adopt binary cross entropy and interaction-over-union as loss functions, but they treats all pixels equally and independently. Since the imbalance distribution between boundary pixels and non-boundary ones, it is difficult to predict clear boundary. To solve this problem, various loss functions \cite{DBLP:conf/aaai/WeiWH20}, \cite{DBLP:journals/tip/ChenZLYX21}, \cite{zhu2021supplement} have been proposed to enhance the boundary details by assigning high weights to the pixels near boundary. Despite impressive, considering only the pixels around boundary is not comprehensive enough, pixels with similar appearance located in the border region between foreground and background are prone to wrong. Therefore, attention should be paid to pixels in grades, rather than treating the pixels inside the object equally.
\par To address above problems, we propose the sharp eyes network (SENet), which based on the strategy of separating first and then segmenting to generate more complete saliency maps with low noises. Compared with the method based on supplementing information, the proposed method is reverse, i.e., the detection range is large to small, which is consistent with the characteristics of the human vision. First, to prevent background interference and locate the complete targets, we design target separation (TS) branch to produce saliency features with expanded boundary (SF w/ EB), which is a fractal structure under the supervision of expanded ground truth encoded by the operations of dilation and erosion. Compared with edge or skeleton information, the proposed SF w/ EB is able to enlarge the differences between salient objects and cluttered background in detail. Second, to get fine salient objects, we develop the object segmentation (OS) branch, which consists of feature aggregation (FA) module and feature interaction (FI) module. The FA module is used to generate more complementary features, which aggregates high-level semantic features, low-level detail features, and global context information in a progressively way. The FI module is applied to improve the structural and detailed information of the salient objects, which feed the SF w/ EB into the above aggregated features by element-wise multiplication. The interactive features share the properties of consistent semantics and structural information. Moreover, to suppress the redundancy and noise caused by interactive features, we use a series of weighted convolution layers to refine them. Finally, the FA module and FI module will cascade three times to generate the final saliency maps. In each cascade, the original ground truth is applied to supervise the output of the FI module to stop the SF w/ EB confusing the model in learning process.
\par Furthermore, we propose a hierarchical difference aware (HDA) loss to further improve the prediction of structural integrity and local details. Different from previous loss functions which assign high weights to pixels around boundary, the proposed HDA loss distributes weights to pixels according to the distance from boundary hierarchically, which assigns each pixel a weight by calculating the sum of differences between a pixel and pixels in its neighborhoods in dilated ground truth and eroded one. In consequence, hard pixels in border region will be given more attention to discriminate the similar parts between foreground and background. In summary, our main contributions can be summarized as follows:
\par 1) We propose a novel salient object detector working the same way of human
visual characteristics, which separates target first and then segment it. It is a framework of detection range from large to small and segmentation accuracy from coarse to fine, which can integrate expanded saliency features and aggregated complementary features to generate more complete but less noisy saliency maps.  
\par 2) We design a hierarchical difference aware loss that assigns weights to pixels according to their distance from boundary in grades, which can further improve the prediction of structural integrity and local details by focusing on the indistinguishable regions.
\par 3) Extensive experimental results on five benchmark datasets demonstrate that
the proposed model outperforms 15 state-of-the-art methods in terms of both quantitative and qualitative evaluations. In addition, our model is more efficient with an inference speed $39.5$ FPS when processing a $320 \times 320$ image.
\section{Related Work}
Large amounts of image-based SOD methods have been proposed in the past decade. Traditional methods mainly produce saliency maps based on intrinsic cues, such as center prior \cite{DBLP:journals/ijcv/WangJYCHZ17}, color difference \cite{DBLP:conf/cvpr/AchantaHES09} and contrast \cite{DBLP:journals/pami/ChengMHTH15}. Although above methods can obtain good saliency maps in most simple cases, they are not robust to some complex scenes due to the lack of high-level semantic information. With the breakthrough of deep learning in computer vision, many methods based on convolution neural network (CNN) have remarkable boosted the performance of SOD. In this section, we briefly review two mainstream CNN-based methods, including methods based on aggregating multi-level features and methods based on introducing auxiliary information (e.g., edge or skeleton).
\subsection{Methods Based on Aggregating Multi-level Features}
To identify the objects with various scales, many methods integrated the semantics and details from feature pyramid by aggregating multi-level features. Among them, Hou et al. \cite{DBLP:journals/pami/HouCHBTT19} introduced short connections to the skip-layer structures for providing more advanced representations at each layer. Zhang et al. \cite{DBLP:conf/cvpr/ZhangDLH018} proposed a gated bi-directional message passing module to adaptively incorporate multi-level features. Deng et al. \cite{DBLP:conf/ijcai/DengHZXQHH18} exploited an iterative strategy to refine the saliency maps by leveraging low-level features and high-level features alternatively. Liu et al. \cite{DBLP:journals/tip/LiuHY20} proposed an attention network to learn the global and local context of each pixel. Chen et al. \cite{DBLP:conf/aaai/ChenXCH20} proposed the context-aware modules  to aggregate low-level appearance features, high-level semantic features, and global context features. Wei et al. \cite{DBLP:conf/aaai/WeiWH20} presented the cross-feature module to eliminate differences between different layers by multiplication and refined multi-level features in a cascade way. Wang et al. \cite{DBLP:conf/aaai/WangCZZ0G20} proposed a effective framework to progressively polish the multi-level features to be more representative. Pang et al. \cite{DBLP:conf/cvpr/PangZZL20} designed the aggregated modules to integrate the features from adjacent levels, and exploited the self-interaction modules to produce multi-scale representations. Ma et al. \cite{DBLP:conf/aaai/MaXL21} aggregate adjacent features in pairs with layer-by-layer shrinkage to capture the effective details and semantics. Zhang \cite{DBLP:conf/mm/ZhangLPYL21} designed an architecture search framework to alleviate human experts’ effort in the manual design of multi-scale features fusion. Li et al. \cite{DBLP:journals/tip/LiSXMT21}  proposed the promotion and rectification attention to purify the error-prone regions. zhuge et al. \cite{zhuge2021salient} aggregated features with various receptive fields to increase feature diversity.
\par Although the above methods have made great progress, limited by the repeated sub-sampling operations, the detailed information of objects with complex structures will be degenerated, which may suffer from incomplete predictions in some complex scenes with cluttered background.
\subsection{Methods Based on Introducing Auxiliary Information}
To complement the local details, many methods assist the network in generating saliency maps by introducing auxiliary information, such as edge and skeleton. These methods are designed to explore the complementarity of diverse information. Liu et al. \cite{DBLP:conf/cvpr/LiuHCFJ19} utilized an extra edge dataset for joint training with saliency dataset to improve the local details of salient objects. Zhao et al. \cite{DBLP:conf/cvpr/ZhaoW19} and Zhao et al. \cite{DBLP:conf/iccv/ZhaoLFCYC19} directly built the edge loss with binary cross entropy to learn more detailed information of boundaries. Feng et al. \cite{DBLP:conf/cvpr/FengLD19}, Wei et al. \cite{DBLP:conf/aaai/WeiWH20} and Chen et al. \cite{DBLP:journals/tip/ChenZLYX21} adopted a edge-enhanced loss to address the problem of blur edges. Su et al. \cite{DBLP:conf/iccv/SuLZXT19}, Zhao et al. \cite{DBLP:conf/mm/ZhaoXXL21} and Zhu et al. \cite{zhu2021supplement} applied a separate branch to predict salient boundaries, which can better complement the boundary details of salient objects. Wu et al. \cite{DBLP:conf/iccv/WuSH19} and Zhou et al. \cite{DBLP:conf/cvpr/ZhouXLCY20} proposed a bi-directionally feature refinement mechanism to transfer complementary information between the two tasks of saliency and edge. Wei et al. \cite{DBLP:conf/cvpr/WeiWWSH020} decomposed the ground truth into body map and detail map, and adopted two branches to concentrate on center areas of objects and edges respectively. Although the information of edge is introduced to framework, they only contain small parts of the complete object, which is difficult to fill the interior of object. To accurately locate the boundaries and the interiors of the objects at the same time, recent methods introduced edge and skeleton information into framework simultaneously. Liu et al. \cite{DBLP:journals/tip/LiuHC20}  developed a dynamic feature integration strategy to explore the feature combinations automatically between salient object, edge and skeleton detection. Wu et al. \cite{DBLP:journals/tip/WuSH21} proposed a cross multi-branch decoder, which iteratively takes advantage of cross-task aggregation to integrate the features between saliency, edge, and skeleton. However, the divergence of different tasks may bring noises and hard to assistant network uniformly highlight entire salient objects.
\par Whether the methods based on aggregating multi-level features or introducing auxiliary information, they are focused on how to supplement the defects of the object, which lead them hard to get complete prediction in complex scenes. Different above methods by enriching information, we propose a framework that utilize the expanded saliency features to guide the network obtain complete saliency maps with low noises.
\begin{figure*}[ht]
	\centering\includegraphics[scale=0.9]{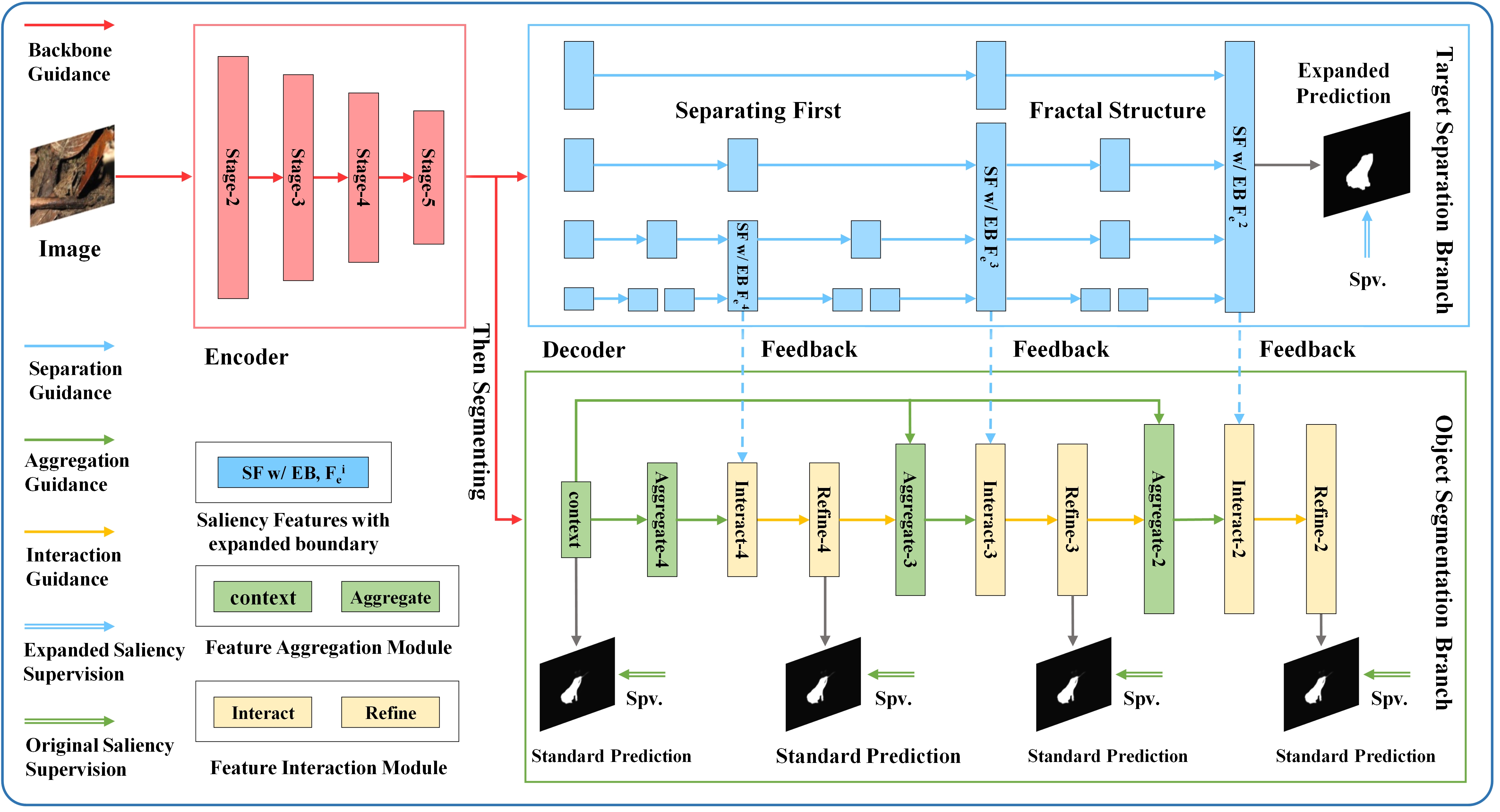}
	\caption{Overview architecture of the proposed model. The encoder is based on ResNet-50, including stage2-stage5. The decoder consists of target separation (TS) branch and object segmentation (OS) branch. Multi-level supervision is applied to optimize the proposed network.}
\end{figure*}
\section{Methodology}
In this section, we first outline the overview architecture of the proposed model. Then, we explicate how each branch works in detail. Finally, we describe the proposed loss function.
\subsection{Network Overview}
As Fig. 2 shows, the proposed network adopts an asymmetric encoder–decoder architecture. The encoder is based on ResNet-50 \cite{DBLP:conf/cvpr/HeZRS16} to extract multi-level features. The decoder mainly consists of target separation (TS) branch and object segmentation (OS) branch. Among them, the TS branch construct a fractal structure to generate saliency features with expanded boundary (SF w/ EB). The OS branch progressively integrates the SF w/ EB and aggregated features to obtain the complete saliency maps. To accelerate the optimization, the proposed hierarchical difference aware loss and multi-level supervision strategy is applied to the network.
\subsection{Feature Encoder}
We utilize ResNet-50 \cite{DBLP:conf/cvpr/HeZRS16} as our backbone network since it achieves a better balance in speed and accuracy. Specifically, since the low-level features bring more computational cost but improvement are limited \cite{DBLP:conf/cvpr/WuSH19}, we only use the last four residual blocks of ResNet-50 to obtain four level features, which are denoted as $\{F_i|i= 2, 3, 4, 5\}$.  Given an input image with size $H \times W \times 3$, the size of the $i$-th feature is $\frac{H}{2^i} \times \frac{W}{2^i} \times C_i$, where $C_i$ is the channel of the $i$-th level feature, and the value of $C_i$ is $64 \times {2^i}$.
\subsection{Target Separation Branch}
To effectively separate the salient objects from background, we proposed the target separation (TS) branch to produce SF w/ EB, which can enlarge the detail differences between foreground and background. As shown in Fig. 2, the TS branch is a self-similar fractal structure under the supervision of the proposed expanded ground truth, which aims to smoothly transfer semantics of high-level features to the low-level ones through different paths. Although this structure is similar in spirit to our previous work \cite{zhu2021supplement}, the TS branch significantly differs from it in two aspects: 1) The supervision signals of the two branches are different (i.e. the features generated by the two branches are different). 2) The sub-modules used in the two branches are different. Since the range of SF w/ EB is larger than the original saliency features, the TS branch adds the module to enlarge the visual receptive field and the attention module to surpress the noise. The generation details of the expanded ground truth and the SF w/ EB are described as follows.
\subsubsection{Expanded Ground Truth Generation}
Considering the boundary is the primary information to distinguish the foreground and background, we construct the expanded ground truth to supervise the output of the TS branch, thus generating the SF w/ EB. Fig. 3 shows the construction process of expanded ground truth and the prediction of different components of the network. As can be seen, compared with the original ground truth, the expanded ground truth contains the expanded object boundary. Under the supervision of the expanded ground truth, the TS branch output the  SF w/ EB. In addition, to reflect the benefits to the network from the SF w/ EB, the prediction of the whole network SENet are shown in the last column.
\par In detail, we first perform dilation and erosion operations on the original ground truth respectively, which are denoted as Dilate (.) and Erode (.). Then, we calculate the difference set of Dilate (.) and Erode (.) to obtain a boundary envelope, which includes both inside and outside of the boundary. Finally, we calculate the union of boundary envelope and original ground truth to generate the expanded ground truth. The boundary envelope and expanded ground truth are defined in Eq. 1 and Eq. 2, respectively:
\begin{equation}
G{T_{be}} = {\mathop{\rm Dilate}\nolimits} {(GT,k)^i} - {\mathop{\rm Erode}\nolimits} {(GT,k)^i},
\end{equation}
\begin{equation}
G{T_e} = G{T_{be}} \cup GT,
\end{equation}
where $GT$ is the original ground truth, $GT_{be}$ indicates the boundary envelope and $GT_{e}$ represents the expanded ground truth, the size of kernel $k$ is $3 \times 3$ and the iteration $i$ is set to 5 empirically, $-$ denotes subtraction operation and  $\cup$ represents union operation. 
\begin{figure}[ht]	
	\subfigure[]{
		\begin{minipage}[b]{0.135\linewidth}				
			\includegraphics[width=1.16\linewidth]{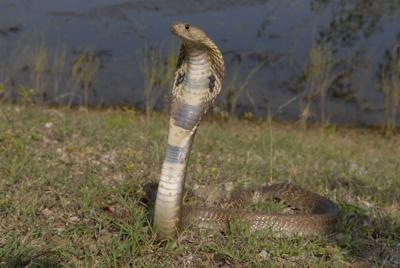}\vspace{1.5pt}	
			\includegraphics[width=1.16\linewidth]{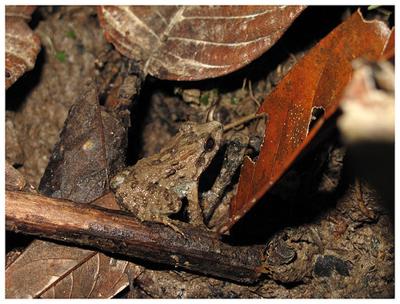}\vspace{1.5pt}
			\includegraphics[width=1.16\linewidth]{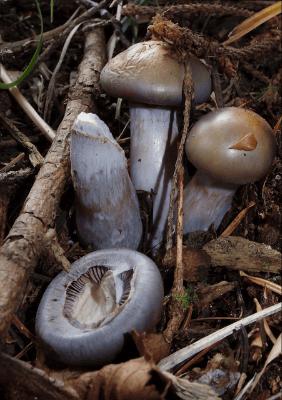}
	\end{minipage}}
	\subfigure[]{
		\begin{minipage}[b]{0.135\linewidth}			
			\includegraphics[width=1.16\linewidth]{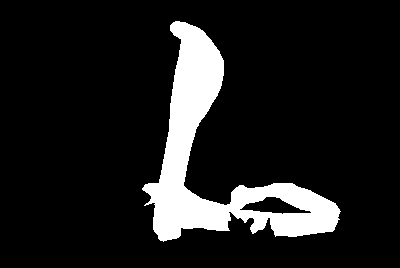}\vspace{1.5pt}	
			\includegraphics[width=1.16\linewidth]{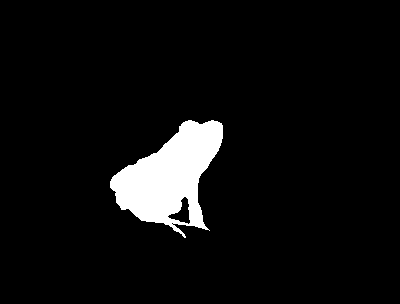}\vspace{1.5pt}
			\includegraphics[width=1.16\linewidth]{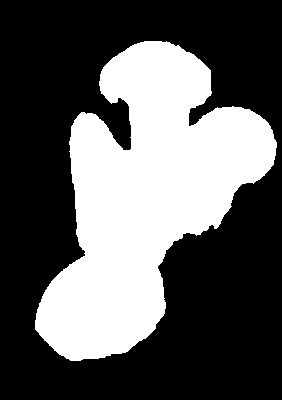}			
	\end{minipage}}
	\subfigure[]{
		\begin{minipage}[b]{0.135\linewidth}
			\includegraphics[width=1.16\linewidth]{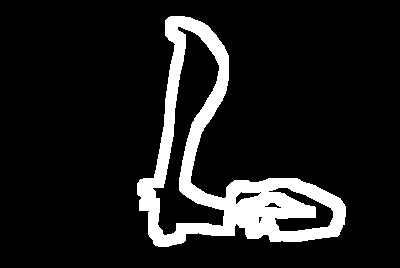}\vspace{1.5pt}	
			\includegraphics[width=1.16\linewidth]{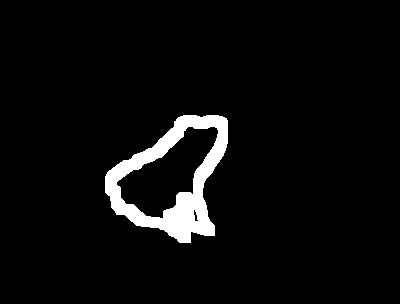}\vspace{1.5pt}	
			\includegraphics[width=1.16\linewidth]{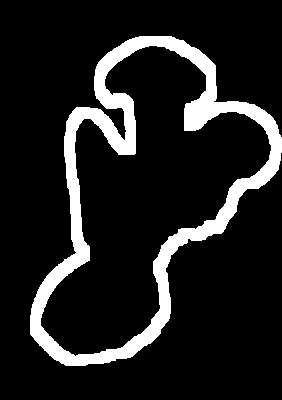}		
	\end{minipage}}
	\subfigure[]{
		\begin{minipage}[b]{0.135\linewidth}
			\includegraphics[width=1.16\linewidth]{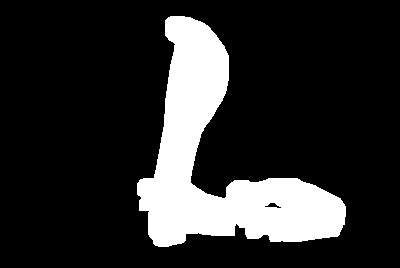}\vspace{1.5pt}	
			\includegraphics[width=1.16\linewidth]{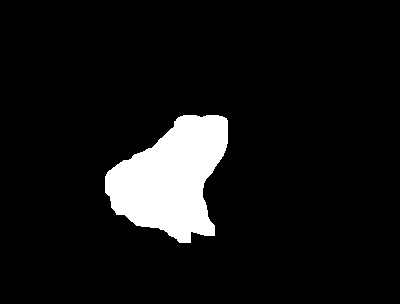}\vspace{1.5pt}	
			\includegraphics[width=1.16\linewidth]{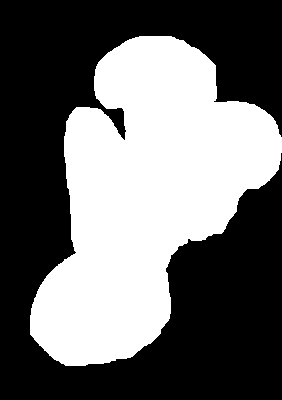}		
	\end{minipage}}
	\subfigure[]{
		\begin{minipage}[b]{0.135\linewidth}
			\includegraphics[width=1.16\linewidth]{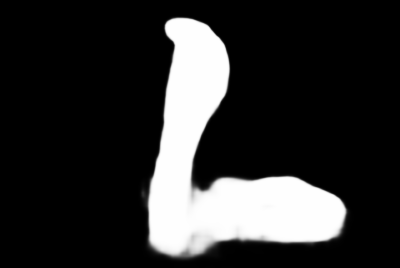}\vspace{1.5pt}	
			\includegraphics[width=1.16\linewidth]{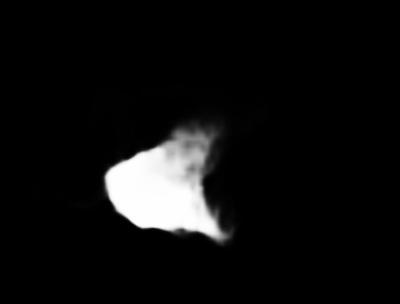}\vspace{1.5pt}
			\includegraphics[width=1.16\linewidth]{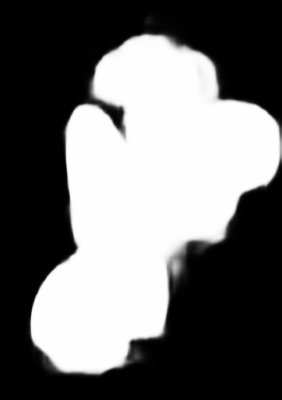}			
	\end{minipage}}
	\subfigure[]{
	\begin{minipage}[b]{0.135\linewidth}
		\includegraphics[width=1.16\linewidth]{fig1/Ours/ILSVRC2012_test_00089997.png}\vspace{1.5pt}	
		\includegraphics[width=1.16\linewidth]{fig1/Ours/ILSVRC2012_test_00061303.png}\vspace{1.5pt}
		\includegraphics[width=1.16\linewidth]{fig1/Ours/6681.png}			
	\end{minipage}}
	\caption{(a) Image. (b) Original ground truth. (c) Boundary envelope. (d) Expanded ground truth. (e) Prediction of TS branch. (f) Prediction of the whole network SENet. We can observe that under the supervision of expanded ground truth, the salient objects predicted by TS branch contains expanded boundaries, which can be used to enlarge the detail difference  between salient object and background and guide our model generate complete saliency maps.}		
\end{figure}
\subsubsection{Saliency features with expanded boundary Generation}
\par The holistic information communication strategy of the proposed TS branch is that high-level features are encoded and injected into low-level features via multiple paths, as shown in Fig.2. Specifically, we first use one $1 \times 1$ convolution layer $conv1$ to the feature maps $\{F_{i}|i= 2, 3, 4, 5\}$ generated by encoder to obtain the compressed feature representation with 256 channels. Then, the ASPP module \cite{DBLP:journals/corr/ChenPSA17} is used to high-level features $F_{4}$ and $F_{5}$ to capture the multi-scale contextual information, which aims to enhance the semantic correlation among high-level internal features. Next, the RFB module \cite{DBLP:conf/eccv/LiuHW18} is applied to low-level features $F_{2}$ and $F_{3}$ to enlarge the receptive fields of them, which aims to learn the global view of the low-level features. Finally, we utilize the fractal structure to repeat the above feature fusion process until $F_{2}$ is fused, which aims to promote the smooth transmission of information and alleviate the degradation of the network. During the fusion process, the TS branch will output three levels of SF w/ EB, which will be embedded into the object segmentation branch to guide the network to generate complete prediction. The process can be described as follows:
\begin{equation}
F_e^4 = Ca(\delta (up(\theta (\eta (con{v_1}({F_5}))) + \eta (con{v_1}{\rm{(}}{F_4}))),
\end{equation}
\begin{equation}
F_e^3 = Ca(\delta (up(\theta (F_e^4) + con{v_1}(F_e^4)){\rm{ + }}\zeta {\rm{(}}con{v_1}{\rm{(}}{F_3})))),
\end{equation}
\begin{equation}
\begin{aligned}
F_e^2 = Ca(\delta (up(\theta (\delta (\theta (F_e^3) + con{v_1}(F_e^3))) + \\
{\rm{}}con{v_1}(\delta (\theta (F_e^3) + con{v_1}(F_e^3))){\rm{ + }}\\
{\rm{}}con{v_1}{\rm{(}}F_e^3)) + {\theta ^*}(\zeta (con{v_1}({F_2})))),
\end{aligned}
\end{equation}
where $\{F^i_{e}|i= 2, 3, 4\}$ represent the three level SF w/ EB generated by the TS branch , the size of which are $\frac{H}{16} \times \frac{W}{16} \times 256$, $\frac{H}{8} \times \frac{W}{8} \times256$ and $\frac{H}{4} \times \frac{W}{4} \times 256$, $conv_1$ represents a $1 \times 1$ convolution layer for channel reduction, $\eta$ is the ASPP module and $\zeta$ denotes RFB module, $\theta$ represent a series convolution layers, including one $1 \times 1$ convolution layer with 64 channels, one $3 \times 3$ convolution layer with 64 channels and one $1 \times 1$ convolution layer with 256 channels, two ReLU functions are located between the above three convolution layers, $theta^*$ is used to encode the processed features of the RFB module, including one $3 \times 3$ convolution layer with 64 channels, one ReLU functions and one $1 \times 1$ convolution layer with 256 channels, $up$ means the bilinear interpolation operation, $\delta$ denotes the ReLU activation function, and $Ca$ indicates the channel attention module \cite{DBLP:conf/cvpr/HuSS18}.
\subsubsection{Loss of target separation branch}
Under the supervision of the proposed expanded ground truth, we use binary cross entropy (BCE) and interaction-over-union (IoU) functions to minimize the loss of TS branch $L_{c}$. In detail, one  $3 \times 3$ convolution layer is applied to squeeze the number of channels of $F^2_{e}$ to 1. Then, the squeezed feature map is up-sampled to the same size as the ground truth by bilinear interpolation, denoted as $F^1_{e}$. Finally, the sigmoid function is used to $F^1_{e}$ to normalize the predicted values into [0,1]. The process can be described as:
\begin{equation}
F_e^1 = up(con{v_3}(F_e^2),
\end{equation}
\begin{equation}
{L_{c}} = {\mathop{\rm Mean}\nolimits} ({\mathop{\rm BCE}\nolimits} (\sigma (F_e^1),G{T_{e}}){\rm{ + }}{\mathop{\rm IoU}\nolimits} (\sigma (F_e^1),G{T_{e}})),
\end{equation}
where $F^2_{e}$ and $F^1_{e}$ are the SF w/ EB produced by TS branch, the size of which are $\frac{H}{4} \times \frac{W}{4} \times 256$ and $H \times W \times 1$. $up$ represents bilinear interpolation operation, $\sigma$ means Sigmoid function, Mean(.) returns the average of the array elements. 
\begin{figure*}[ht]
	\includegraphics[scale=0.98]{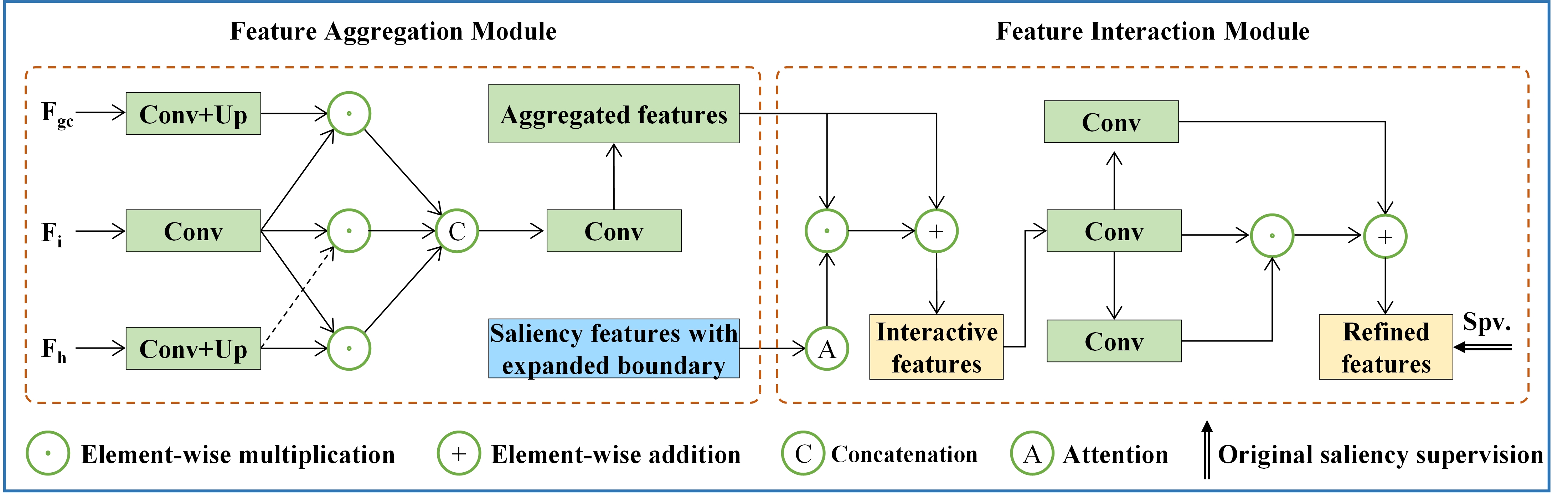}
	\caption{The cascade process of feature aggregation module and feature interaction module in once.}
\end{figure*}
\subsection{Object Segmentation Branch}
To selectively integrate multi-level features and SE w/ EB, we propose object segmentation (OS) branch, which includes feature aggregation (FA) module and feature interaction (FI) module. First, FA module is used to aggregate the multi-level features from encoder and decoder to generate complementary features. Then, FI module adopt a feedback mechanism, where the SE w/ EB produced by TS branch will be fed into the above aggregated features to recover more structural and detailed information. Next, to suppress the redundancy contained in above interactive features, we utilize a series of convolution layers[2] to refine them. Finally, we progressively cascade above three steps in three times to learn more discriminative features and generate final saliency maps. In each cascade, the original ground truth will be used to supervise the output of the FI module, thus guiding the network learn the standand saliency features. Fig. 4 shows the cascade process of FA module and FI module in once.
\subsubsection{Feature Aggregation Module}
Considering that features at different levels contain different attributes, the FA module aims to obtain complementary feature representation by aggregating semantics of high-level features, details of low-level features and global context features. The global context features are used to infer the correlation between different salient regions from the global view of the image. Therefore, we can encode the top layer $F_{5}$ with large receptive field to generate global context features. Specifically, to make the followed module carry out in a lower dimensional space, we first apply one $1 \times 1$ convolution layer to the top layer $F_{5}$ to obtain a compressed feature representation with 256 channels. Then, an ASPP module is applied to above compressed features to capture the multi-scale contextual information, denoted as $F^5_{h}$. Finally, to restore information of original top layer, we apply one $1 \times 1$ convolution layer to $F^5_{h}$ to obtain the features with 2048 channels, and assign the context weight to them in a residual connection way from the dimensions of space and channel, thus suppressing the redundancy and noise caused by the increase channels and learning the discriminative features. The process can be described as:
\begin{equation}
F_h^5 = \eta (\delta (con{v_1}{\rm{(}}{F_5}))),
\end{equation}
\begin{equation}
\begin{aligned}
{\omega _s} = \sigma (bn(con{v_1}(F_h^5))),\\
F_s^5 = {\omega _s} \odot {F_5} + {F_5},
\end{aligned}
\end{equation}
\begin{equation}
\begin{aligned}
{\omega _c}{\rm{ = }}\sigma (con{v_1}(\delta (con{v_1}(gap(F_s^5))))),\\
F_{gc} = {\omega _c} \odot \delta (bn(con{v_1}(F_s^5))),
\end{aligned}
\end{equation}
where $F^5_{h}$ is the multi-scale context features processed by ASPP module, $\omega_s$ and $F_s^5$ denote the spacial weights and weighted features, $\omega_c$ and $F_{gc}$ denote the channel weights and the final global context features, $conv_{1}$ denotes $1 \times 1$ convolution layer, $\eta$ is the ASPP module, $bn$ denotes the batch normalization layer, $\sigma$ means the Sigmoid function, $\odot$ means element-wise multiplication, $gap$ represents the global average pooling and $\delta$ denotes the ReLU activation function.
\par Based on the above global context features, we can aggregate them with high-level and low-level features. In detail, as shown in Fig. 4, the FA module first receives three inputs, including the high level features $\{F^i_{h}|i=5,4,3\}$ output by previous stage decoding, low level features $\{F_{i}|i=4,3,2\}$ from the backbone, and the global context features $F_{gc}$. Note that there are two forms of high-level features, one is the multi-scale context features $\{F^i_{h}|i=5\}$, and the other is $\{F^i_{h}|i=4,3\}$ generated by the feature interaction module to be described below. In other words, the aggregated features output by the FA module will be input into the FI module to interact with the saliency features with expanded boudary. The interactive features output by the FI module will be aggregated into the FA module of the next stage as high-level features. Next, the above three parts are concatenated through a series of convolution and up-sampling operations to obtain the aggregation features $\{F^i_{fa}|i=4,3,2\}$, which are the final output of the FA module:
\begin{equation}
\begin{aligned}
F_{fa}^i = conv(cat(conv(F_i) \odot up(conv({F_{gc}})),\\
conv(F_i) \odot up(conv(F_h^{i+1}))),\\
conv(F_i)) \odot conv(F_h^{i+1}))) \enspace {\rm{  }} i=4,3,2, 
\end{aligned}
\end{equation}
where $conv$ indicates one $3 \times 3$ convolution layer followed by a batch normalization layer and ReLU activation function, $\odot$ means element-wise multiplication, $up$ denotes bilinear interpolation operation and $cat$ means concatenation operation.
\subsubsection{Feature Interaction Module}
To generate complete and fine saliency features, we propose the FI module, which feed the saliency features with expanded boundary (SF w/ EB) into the aggregated features generated by FA module. Actually, due to the multiple sub-sampling operations, the details in some regions of the image will be lost in the forward propagation process of the network. Even if the FA module is used to generate complementary features, it is still difficult to obtain complete salient objects in a cluttered background, especially the scene with many distracts and disorderly. Considering the SF w/ EB can enlarge the differences of details between salient objects and background, we interact them with aggregated features to help the network get complete prediction. Moreover, to suppress the redundancy and noise caused by interactive features, such as unnecessary background information, we apply a set of weighted convolution layers to them. It is worth noting that the original ground truth will be leveraged to supervise the output of the FI module, so there is no need to worry that SF w/ EB confusing the model in learning process.
\par Specifically, we first adopt coordinate attention \cite{DBLP:conf/cvpr/HouZF21} to the SF w/ EB to captured spatial direction and positional information, which can augment the representations of the objects of interest and suppress noise. Then, we fed the weighted SF w/ EB into the aggregated features by element-wise multiplication with a residual connection. The above interactive features share the properties of consistent semantic and structure information. Next, motivated by the self refinement module \cite{DBLP:conf/aaai/ChenXCH20}, we utilize a series weighted convolution layers to refine the interactive features. Finally, the FA module and FI module are cascaded three times to generate three levels of feature maps $\{F^i_{h}|i=4,3,2\}$, and then one $3 \times 3$ convolution layer, bilinear interpolation and  Sigmoid function are used to multi-scale context features $\{F^i_{h}|i=5\}$ and $\{F^i_{h}|i=4,3,2\}$ to predict the saliency maps $\{P_{i}|i=5,4,3,2\}$, which are the final outputs of the OS branch. The process can be described as: 
\begin{equation}
F_i^i = coa(conv(F_e^i)) \odot F_{fa}^i + F_{fa}^i{\rm{ }} \enspace {\rm{ }} i=4,3,2,
\end{equation}
\begin{equation}
\begin{aligned}
F_h^i = \delta (con{v_3}(conv(F_i^i)) \odot conv(F_i^i) +\\
con{v_3}(conv(F_i^i)){\rm{ }} \enspace {\rm{ }} \enspace {\rm{ }} i=4,3,2,
\end{aligned}
\end{equation}
\begin{equation}
{P_i} = \sigma (up(con{v_3}(F_h^i))  \enspace {\rm{ }} i=5,4,3,2,
\end{equation}
where $F^i_{e}$ represents the SF w/ EB, $F^i_{i}$ denotes the interactive features, $F^i_{h}$ indicates the refined features, i.e. the high-level feature described in above mentioned FA module, coa means the coordinate attention, $conv3$ denotes one $3 \times 3$ convolution layer, $conv$ indicates one $3 \times 3$ convolution layer followed by a batch normalization layer and ReLU activation function, $\odot$ means element-wise multiplication, $\delta$ is ReLU activation function and $\sigma$ represents Sigmoid function. 
\subsection{Hierarchical Difference Aware Loss}
To further improve the completeness of salient object, we propose the Hierarchical difference aware (HDA) loss to supervise the outputs of OS branch. In SOD, binary cross entropy (BCE) and interaction-over-union (IoU) are most widely used loss function. However, these functions treat each pixel independently and equally, which may suffer from vague boundary and incomplete prediction due to the imbalance distribution between boundary and non-boundary. Inspired by the weighted pixel position loss \cite{DBLP:conf/aaai/WeiWH20}, we propose the HDA loss to guide the network focus more on the border region between foreground and background, which can alleviate the unbalanced distribution of boundary pixels and non-boundary pixels. Different from previous loss functions which assign high weights to pixels around boundary, the proposed HDA loss assigns weights to the pixels according to the distance from boundary in grades, i.e. pixels close to the boundary will be assigned high weights, pixels in the border region will be assigned medium weights, and pixels far away from the boundary will be assigned low weights, as shown in Fig. 5. Consequently, pixels with similar appearance in border region will be given more attention, which not only helps to improve the boundary details, but also benefit the integrity of global structure.
\begin{figure}[tp]	
	\centering
	\makeatletter
	\renewcommand{\@thesubfigure}{\hskip\subfiglabelskip}
	\makeatother	
	\subfigure[Image]{
		\begin{minipage}[b]{0.2\linewidth}			
			\includegraphics[width=1.2\linewidth,height=0.8\linewidth]{fig1/Image/ILSVRC2012_test_00089997.jpg}\vspace{1pt}
			\includegraphics[width=1.2\linewidth,height=0.95\linewidth]{fig1/Image/ILSVRC2012_test_00061303.jpg}\vspace{1pt}	
			\includegraphics[width=1.2\linewidth,height=1.7\linewidth]{fig1/Image/6681.jpg}
			
	\end{minipage}}
	\hspace{0.02in}
	\subfigure[GT]{
		\begin{minipage}[b]{0.2\linewidth}
			\includegraphics[width=1.2\linewidth,height=0.8\linewidth]{fig1/Mask/ILSVRC2012_test_00089997.png}\vspace{1pt}
			\includegraphics[width=1.2\linewidth,height=0.95\linewidth]{fig1/Mask/ILSVRC2012_test_00061303.png}\vspace{1pt}
			\includegraphics[width=1.2\linewidth,height=1.7\linewidth]{fig1/Mask/6681.png}
	\end{minipage}}
	\hspace{0.02in}	
	\subfigure[$\omega$]{
		\begin{minipage}[b]{0.2\linewidth}
			\includegraphics[width=1.2\linewidth,height=0.8\linewidth]{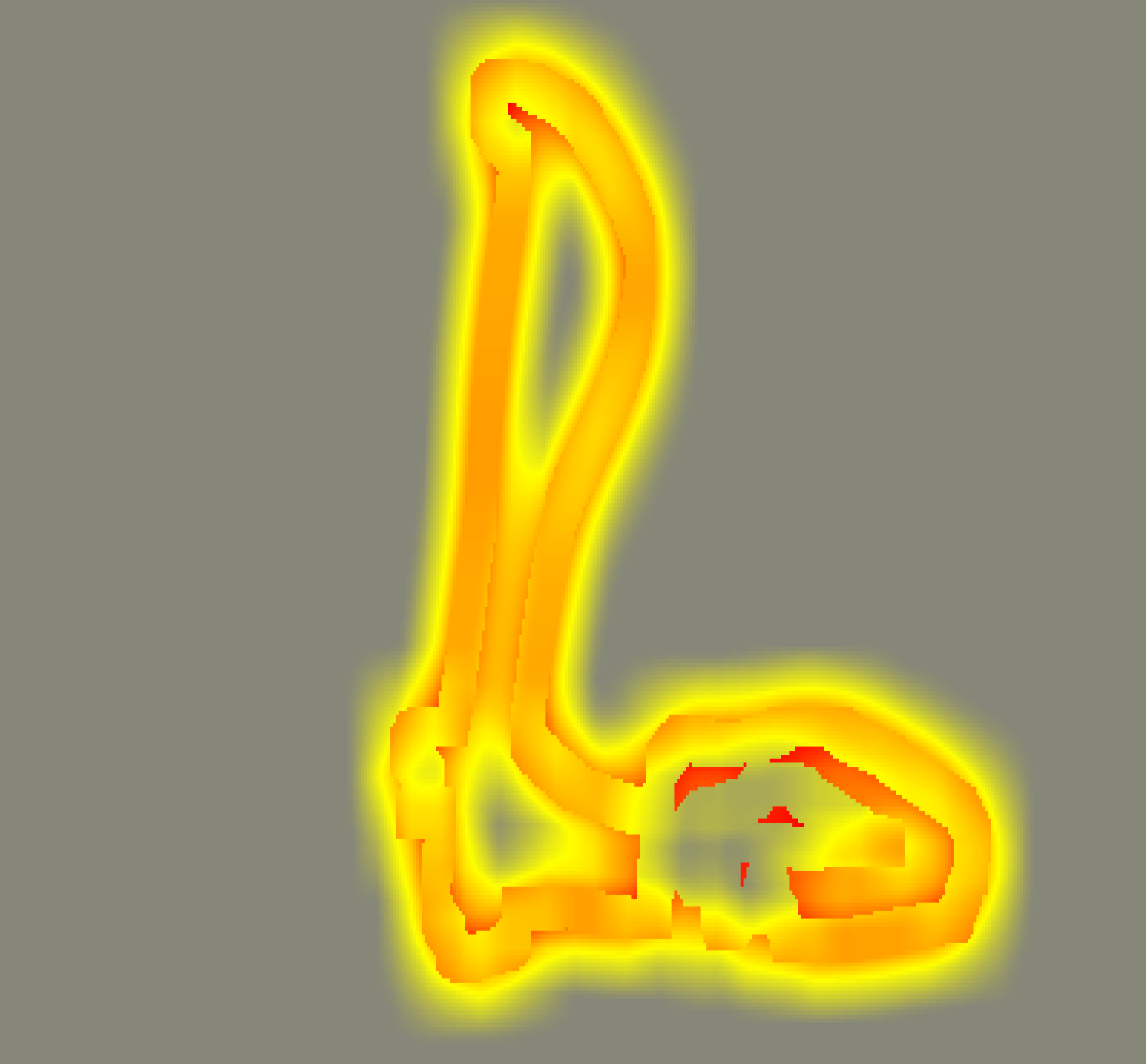}\vspace{1pt}
			\includegraphics[width=1.2\linewidth,height=0.95\linewidth]{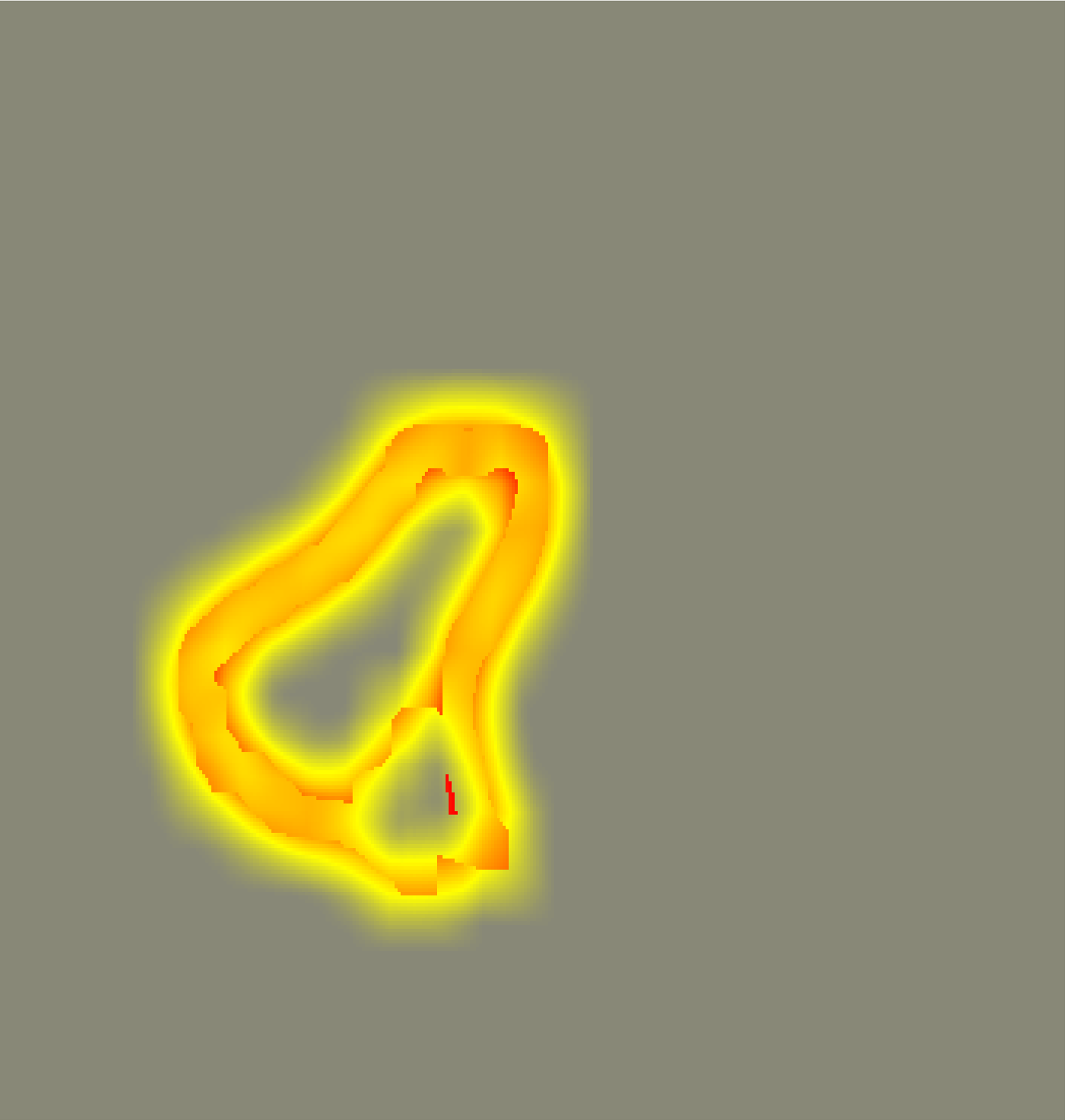}\vspace{1pt}
			\includegraphics[width=1.2\linewidth,height=1.7\linewidth]{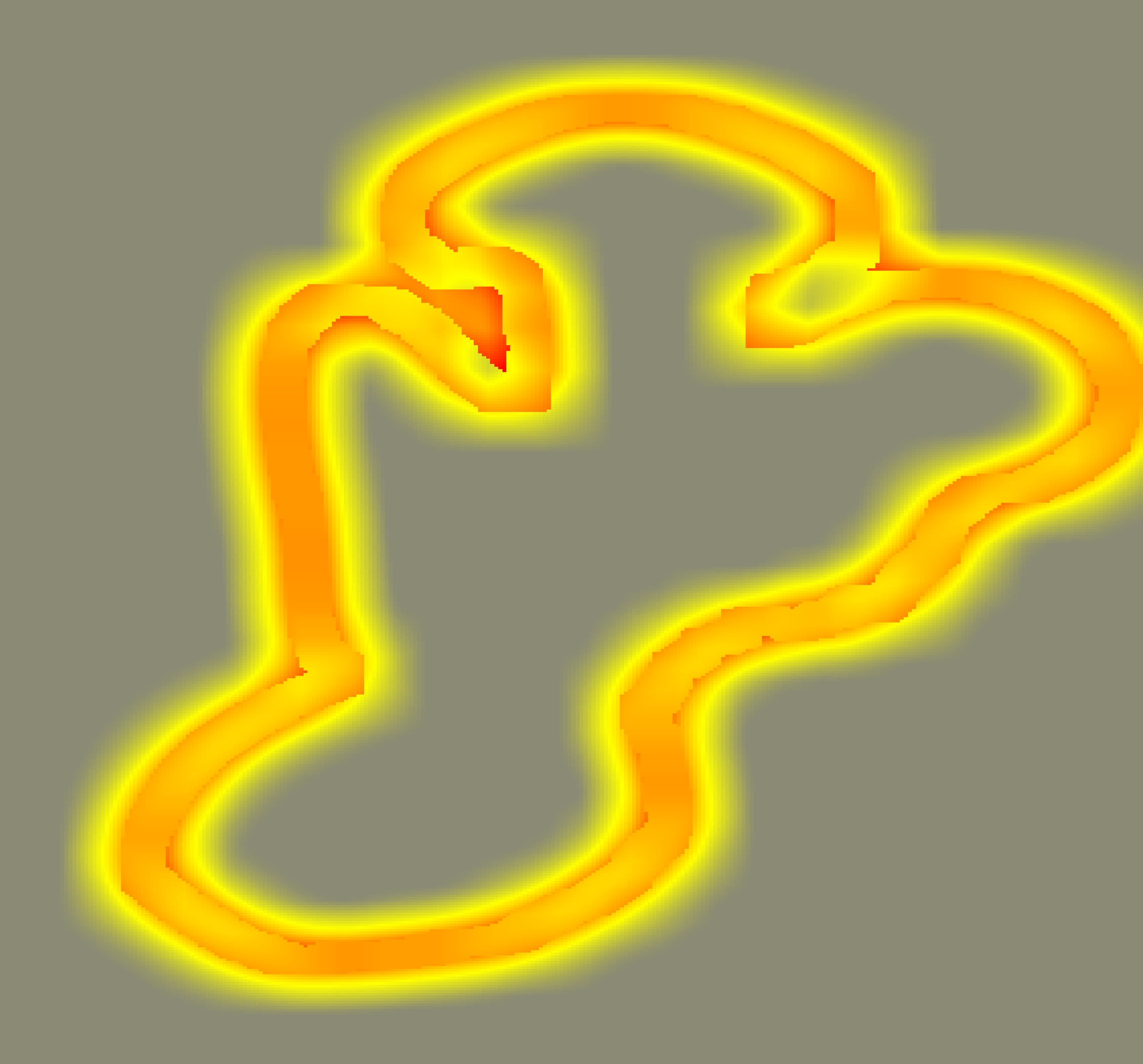}		
	\end{minipage}}
	\hspace{0.02in}
	\subfigure[]{
		\begin{minipage}[b]{0.2\linewidth}			
			\includegraphics[width=0.4\linewidth,height=3.5\linewidth]{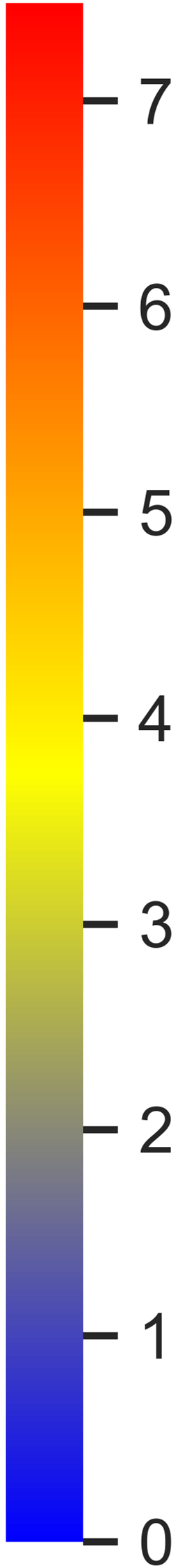}			
	\end{minipage}}
	\caption{Some visual examples of the distribution of $\omega$. We observe that the pixels in border region between salient objects and background are given more attention, which can guide the network identify the indistinguish regions and further improve the completeness of prediction.}
\end{figure} 
\par Specifically, the proposed HDA loss assigns each pixel a weight $\omega_{ij}$ in BCE and IoU loss functions. First, we apply the operations of dilation and erosion to the original ground truth (GT) to obtain the dilated GT and eroded GT. Then, two kinds of weights are calculated based on the difference between a pixel and pixels in its neighborhoods in dilated GT and eroded GT respectively, which represent the importance of each pixel. If a pixel is more different from its surroundings in dilated/eroded GT, it will be assigned a higher weight. Next, we sum the above two weights to further emphasize the importance of pixels close to the boundary. Because there is an intersection between their border regions, the pixels in the intersection of two border regions are calculated two times, which will be assigned higher weights. The  $\omega_{ij}$ are defined in Eq. 15, Eq. 16 and Eq. 17. 
\begin{equation}
\omega _{ij}^d = \left| {\sum\limits_{x,y \in {R_{ij}}} {\left[ {{\mathop{\rm Dilate}\nolimits} {{(G{T_{xy}},k)}^n}} \right] - {\mathop{\rm Dilate}\nolimits} {{(G{T_{ij}},k)}^n}} } \right|,
\end{equation}
\begin{equation}
\omega _{ij}^e = \left| {\sum\limits_{x,y \in {R_{ij}}} {\left[ {{\mathop{\rm Erode}\nolimits} {{(G{T_{xy}},k)}^n}} \right] - {\mathop{\rm Erode}\nolimits} {{(G{T_{ij}},k)}^n}} } \right|,
\end{equation}
\begin{equation}
{\omega _{ij}} = \omega _{ij}^d + \omega _{ij}^e,
\end{equation}
where $\omega_{ij}$ is the sum of differences of pixel position in dilated GT and in eroded GT, i.e., $\omega^d_{ij}$ and $\omega^e_{ij}$. $R_{ij}$ denotes the region that surrounds the pixel ($i$, $j$) with a size of $30 \times 30$. Dilate(.) and Erode (.) represent operations of dilation and erosion, and the size of kernel $k$ is $3 \times 3$ and the iteration $n$ is set to 5 empirically. Based on the proposed $\omega_{ij}$, the weighted BCE and IoU loss functions can be defined as:
\begin{equation}
\begin{aligned}
L_{bce}^\omega  = \frac{{ - \sum\limits_i^H {\sum\limits_j^W {\left[ \begin{array}{l}
				(1 + \lambda  \cdot {\omega _{ij}}) \cdot ((G{T_{ij}} \cdot {P_{ij}}) + \\
				(1 - G{T_{ij}}) \cdot \log (1 - {P_{ij}}))
				\end{array} \right]} } }}{{\sum\limits_i^H {\sum\limits_j^W {\left[ {(1 + \lambda  \cdot {\omega _{ij}})} \right]} } }},
\end{aligned}
\end{equation}
\begin{equation}
L_{iou}^\omega  = 1 - \frac{{\sum\limits_{i = 1}^H {\sum\limits_{j = 1}^W {(1 + \lambda  \cdot {\omega _{ij}}) \cdot (G{T_{ij}}} }  \cdot {P_{ij}})}}{{\sum\limits_{i = 1}^H {\sum\limits_{j = 1}^W {(1 + \lambda  \cdot {\omega _{ij}}) \cdot (G{T_{ij}}} }  + {P_{ij}} - G{T_{ij}} \cdot {P_{ij}})}},
\end{equation}
where $\lambda$ is set to 5 to bias towards the final result. Based on above definition, the proposed hierarchical difference aware loss $L_{hda}$ are shown in Eq. 20, which considers both local and global structure.
\begin{equation}
{L_{hda}} = {\mathop{\rm Mean}\nolimits} (L_{bce}^\omega  + L_{iou}^\omega ),
\end{equation}
where Mean(.) returns the average of the array elements. We use multi-level supervision is applied to optimize the OS branch by the proposed loss, which is the sum of the four losses between predictions $\{P_{i}|i=2, 3, 4, 5\}$ and $GT$, as shown in Eq. 21. 
\begin{equation}
{L_f} = \sum\limits_{i = 2}^5 {{\alpha _i}}  \cdot {L_{hda}}({P_i},GT),
\end{equation}
where $\alpha_{i}$ indicates the weight of different losses, which is set to 1, 0.8, 0.6 and 0.4 from $\alpha_{2}$ to $\alpha_{5}$, respectively. Because the high-level features of network have larger error, they should have smaller weights empirically. The whole loss $L$ is minimized in end-to-end, which consists of both loss of TS branch and loss of OS branch. 
\begin{equation}
L = {L_c} + \beta  \cdot {L_f},
\end{equation}
where $\beta$  is set to 1 to make the above two losses range comparable. 
\begin{table*}[h]
	\centering
	\setlength{\abovecaptionskip}{0pt}
	\setlength{\belowcaptionskip}{3pt}
	\renewcommand\arraystretch{1.4}
	\setlength\tabcolsep{5.35pt}
	\caption{Performance comparisons with 14 state-of-the-art methods over five datasets. Max F-measure ($F_\beta$, larger is better), S-measure ($S_m$, larger is better) and MAE (smaller is better) are used to measure the model performance. The best three results are marked in \textcolor{red}{red} and \textcolor{green}{green}.}	
	\begin{tabular}{l|ccc|ccc|ccc|ccc|ccc}
		\hline
		\multicolumn{1}{l|}{\multirow{2}{*}{Method}} & \multicolumn{3}{|c|}{HKU-IS (No. 4447)}                                          & \multicolumn{3}{c|}{DUTS-TE (No. 5019)}                                         & 
		\multicolumn{3}{c|}{DUT-OMRON (No. 5168)} & 
		\multicolumn{3}{c|}{PASCAL-S (No. 850)}                 & 
		\multicolumn{3}{c}{ECSSD (No. 1000)} \\ \cline{2-16} 
		\multicolumn{1}{c|}{}                        & 
		\multicolumn{1}{c} {$F_\beta$} & \multicolumn{1}{c}{$S_m$} &  $\rm MAE$ & \multicolumn{1}{|c}{$F_\beta$} & \multicolumn{1}{c}{$S_m$} &  $\rm MAE$ &  $F_\beta$   &  $S_m$  &  $\rm MAE$ &  \multicolumn{1}{c} {$F_\beta$} &  $S_m$ &  $\rm MAE$ &  $F_\beta$ &  $S_m$  & $\rm MAE$  \\ \hline
		CPD-R \cite{DBLP:conf/cvpr/WuSH19}                               & 0.925                          & 0.906                      & 0.034 & 0.865                          & 0.869                      & 0.043 &
		0.797                          & 0.825 						& 0.057 &  0.864                          & 0.842 						& 0.072 & 
		0.939      						& 0.918  					& 0.037 
		\\ \hline
		BASNet \cite{DBLP:conf/cvpr/QinZHGDJ19}                            & 
		0.928                          & 0.909                      & 0.032 & 	 
		0.859     					 & 0.866  						& 0.048 & 
		0.805                          & 0.836 						& 0.057 &
		0.857       				  & 0.832  						& 0.076  & 
		0.943                          & 0.916                      & 0.037 
		\\ \hline
		PoolNet \cite{DBLP:conf/cvpr/LiuHCFJ19}                           & 
		0.933                          & 0.917                      & 0.032 &	  	
		0.880       					& 0.883  					& 0.040 &
		0.808                          & 0.836 						& 0.056 & 
		0.869       					& 0.845 					 & 0.074 &  
		0.944                          & 0.921                      & 0.039 
		\\ \hline
		SIBI \cite{DBLP:conf/iccv/SuLZXT19}                                 &  0.931                          & 0.913                      & 0.032 & 		
		0.872     					  & 0.879 						 & 0.040 &
		0.803                          & 0.832 						& 0.059 &
		0.870        					& 0.848  					& 0.070   & 
		0.946                          & 0.924                      & 0.035 
		\\ \hline
		CAG-R \cite{DBLP:journals/pr/MohammadiNBMH20}                               &  0.926                          & 0.904                      & 0.030  & 
		0.866      						& 0.864  					& 0.040 &
		0.791                          & 0.815 						& 0.054& 
		0.866       					& 0.836 					 & 0.066 &
		0.937                          & 0.908                      & 0.037 
		\\ \hline
		GateNet \cite{DBLP:conf/eccv/ZhaoPZLZ20}                              &  0.934                          & 0.915                      & 0.033 & 
		0.887 						& 0.885  						& 0.040 &
		0.818                          & 0.838 						& 0.055 &
		0.945                          & 0.920                       & 0.040  &
		0.875       					& 0.852  					& 0.068  
		\\ \hline
		ITSD \cite{DBLP:conf/cvpr/ZhouXLCY20}                                 &  0.934                          & 0.917                      & 0.031 &  
		0.883     						 & 0.885 					 & 0.041 & 
		0.821       & 0.840  						 & 0.061 & 
		0.876       				& 0.856   					& 0.064   & 
		0.947                          & 0.925                      & 0.035 
		\\ \hline
		MINet \cite{DBLP:conf/cvpr/PangZZL20}                                &  0.935                          & 0.919                      & 0.029  &    0.884      						& 0.884  					& 0.037 & 
		0.810                           & 0.833 					& 0.055  &
		0.873      						 & 0.851  					& 0.064 &
		0.948        & 0.925                      & 0.034  
		\\ \hline
		GCPA \cite{DBLP:conf/aaai/ChenXCH20}                               &  0.938         & 0.920   & 0.031 & 
		0.888      					 & \textcolor{green}{0.891}   & 0.038 &		
		0.812      						& 0.839 						& 0.056 &		
		0.876       				& \textcolor{red}{0.861}  	    & \textcolor{red}{0.061} &      
		0.949           & 0.927   & 0.035   			 
		\\ \hline
		F3Net \cite{DBLP:conf/aaai/WeiWH20}                               & 0.937                          & 0.917                      & \textcolor{green}{0.028} & 0.891            & 0.888                     & 0.036  & 
		0.813       				& 0.839 						 & 0.053 & 0.878        & 0.855 						& \textcolor{green}{0.062} & 
		0.945      						& 0.924  					& \textcolor{green}{0.033}   
		\\ \hline
		PurNet \cite{DBLP:journals/tip/LiSXMT21}                              & 0.935                          & 0.915                       & 0.031  & 0.878                          & 0.881                      & 0.039 & 
		0.814      						 & 0.841 					 & \textcolor{green}{0.051} & 0.873                          & 0.843						 & 0.069 & 0.945     					 & 0.925  							& 0.035  
		\\ \hline
		MSFNet \cite{DBLP:conf/mm/ZhangLPYL21}                                
		& 0.927       				  & 0.907              & \textcolor{red}{0.027}  & 0.877                          & 0.875                      & \textcolor{red}{0.034}  & 0.799       					& 0.820  					& \textcolor{red}{0.046}   & 0.870                           & 0.845 				& \textcolor{red}{0.061}  & 0.941     					 & 0.914 					 & \textcolor{green}{0.033} 
		\\ \hline
		GFINet \cite{zhu2021supplement}                               & \textcolor{green}{0.939}        & \textcolor{green}{0.921}     & \textcolor{green}{0.028} & 0.890     						 & 0.889    & 0.038 &
		0.823       & 0.843  & 0.054 & 0.877          & 0.856 			& 0.064 & 
		0.948       				& \textcolor{red}{0.929}   & \textcolor{red}{0.032}   
		\\ \hline
		DCN \cite{DBLP:journals/tip/WuSH21}                                 & \textcolor{green}{0.939}                          & \textcolor{red}{0.922}                     & \textcolor{red}{0.027} & \textcolor{green}{0.894}                          & \textcolor{green}{0.891}                      & \textcolor{green}{0.035} & 0.823       & \textcolor{green}{0.845}   & \textcolor{green}{0.051}   & 0.878        & 0.857   & \textcolor{green}{0.062} & 
		\textcolor{red}{0.952}     				& \textcolor{green}{0.928}  						& \textcolor{red}{0.032} 
		\\ \hline
		ICON \cite{zhuge2021salient}                                 & \textcolor{green}{0.939}                          & 0.920                     & 0.029 & \textcolor{green}{0.892}                          & 0.889                      & 0.037 & \textcolor{green}{0.825}       & 0.844   & 0.057   & \textcolor{green}{0.881}        & 0.857   & 0.063 & 
		\textcolor{green}{0.950}     				& \textcolor{red}{0.929}  						& \textcolor{red}{0.032}
	    \\ \hline
		\textbf{SENet (Ours)} & \textcolor{red}{0.941}   & \textcolor{red}{0.922}                       & \textcolor{red}{0.027}  & \textcolor{red}{0.899}                           & \textcolor{red}{0.896}                       & \textcolor{red}{0.034}  & \textcolor{red}{0.827}        & \textcolor{red}{0.847}   & \textcolor{green}{0.051}   & \textcolor{red}{0.882}                           & \textcolor{green}{0.859}  & \textcolor{green}{0.062}  & \textcolor{green}{0.950}        & \textcolor{red}{0.929}   & \textcolor{red}{0.032}   \\ \hline
	\end{tabular}
\end{table*} 
\begin{figure*}[ht]
	\centering\includegraphics[scale=.08]{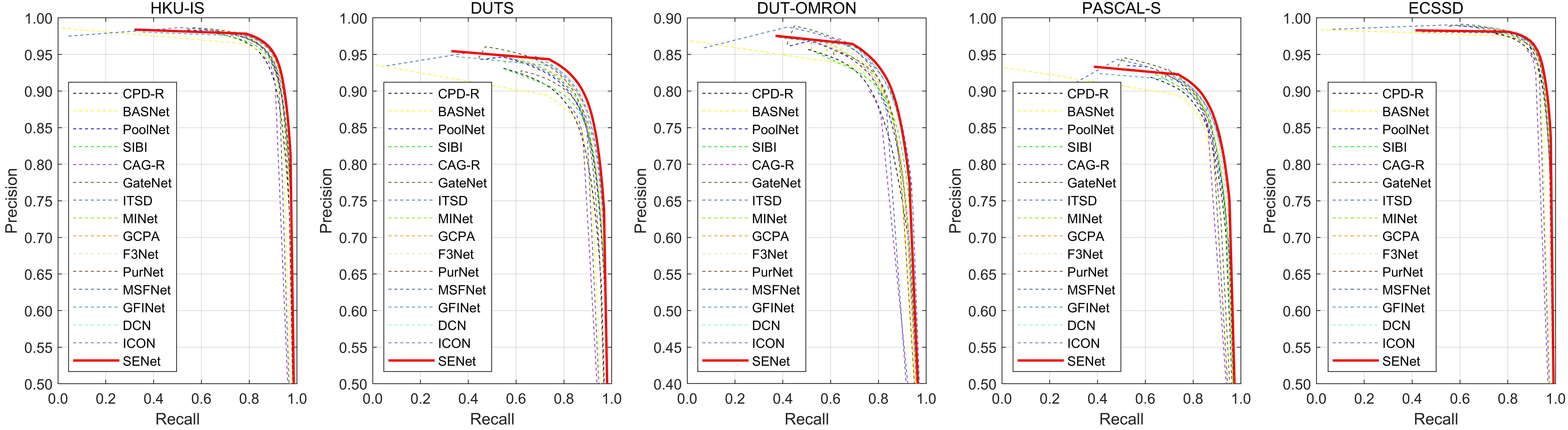}
	\centering\includegraphics[scale=.08]{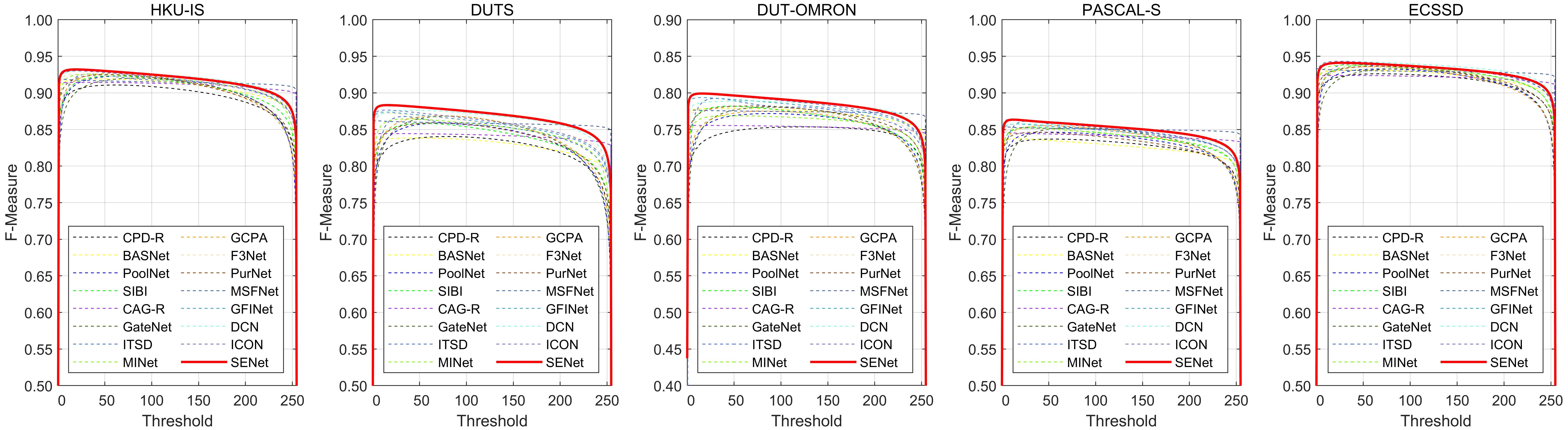}
	\caption{Performance comparisons with 15 state-of-the-art methods over five datasets. The first row shows comparison of precision-recall curves, and the second row shows comparison of F-measure curves over different thresholds.}
\end{figure*}
\begin{figure*}[ht]	
	\centering
	\makeatletter
	\renewcommand{\@thesubfigure}{\hskip\subfiglabelskip}
	\makeatother
	\subfigure[\tiny Image]{
		\begin{minipage}[b]{0.068\linewidth}			
			\includegraphics[width=1.15\linewidth]{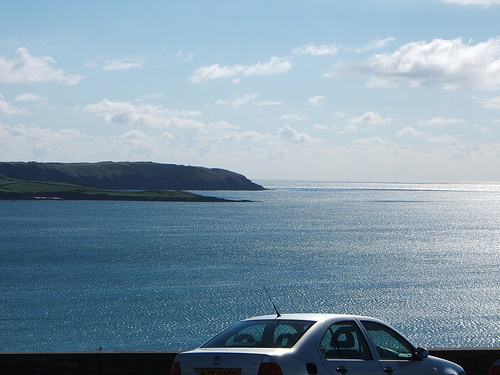}\vspace{1.5pt}
			\includegraphics[width=1.15\linewidth]{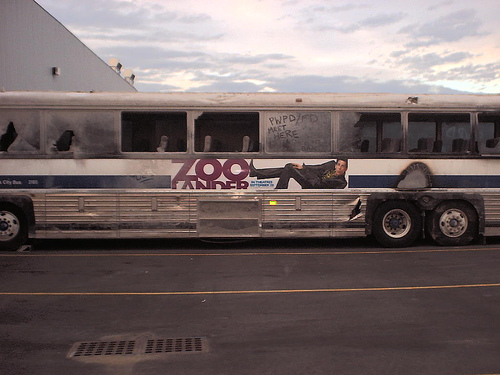}\vspace{1.5pt}
			\includegraphics[width=1.15\linewidth]{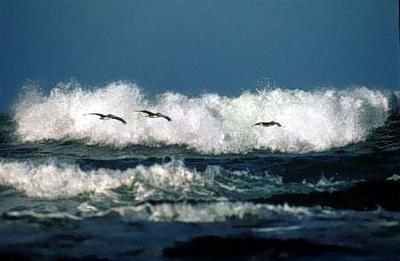}\vspace{1.5pt}
			\includegraphics[width=1.15\linewidth]{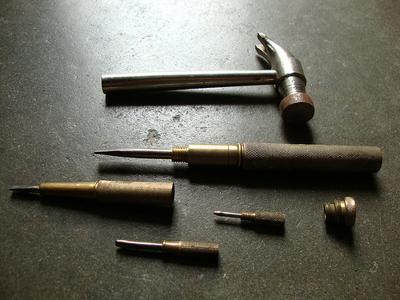}\vspace{1.5pt}
			\includegraphics[width=1.15\linewidth]{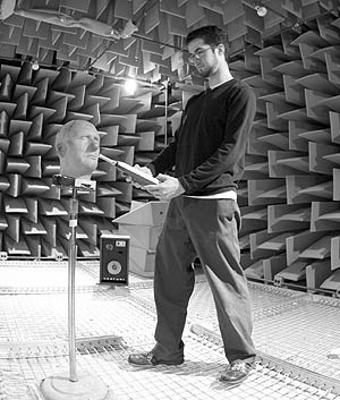}\vspace{1.5pt}
			\includegraphics[width=1.15\linewidth]{fig1/Image/ILSVRC2012_test_00061303.jpg}\vspace{1.5pt}
			\includegraphics[width=1.15\linewidth]{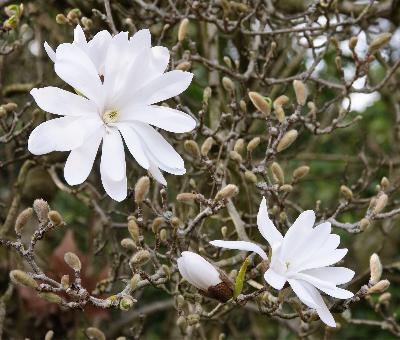}\vspace{1.5pt}
			\includegraphics[width=1.15\linewidth]{fig1/Image/6681.jpg}\vspace{1.5pt}
			\includegraphics[width=1.15\linewidth]{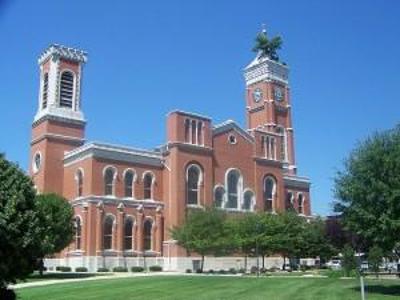}\vspace{1.5pt}
			\includegraphics[width=1.15\linewidth]{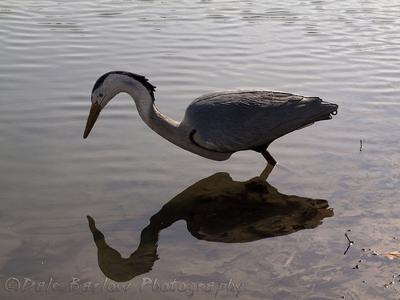}\vspace{1.5pt}
			\includegraphics[width=1.15\linewidth]{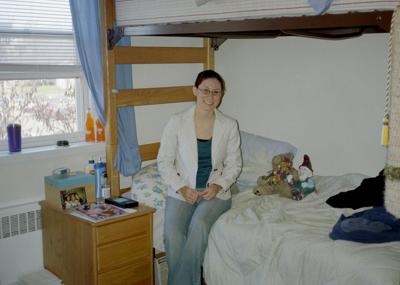}				
	\end{minipage}}
	\subfigure[\tiny GT]{
		\begin{minipage}[b]{0.068\linewidth}			
			\includegraphics[width=1.15\linewidth]{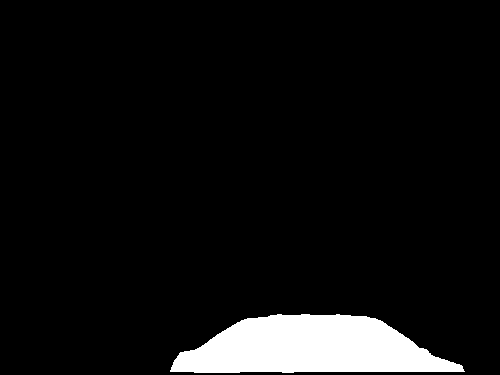}\vspace{1.5pt}
			\includegraphics[width=1.15\linewidth]{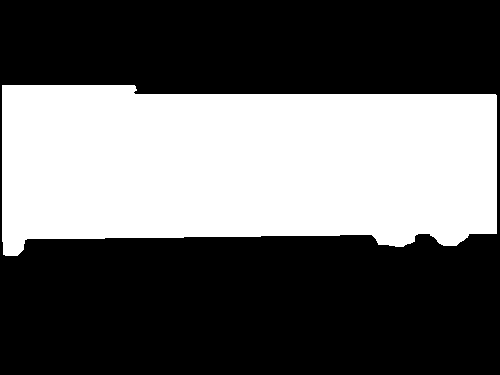}\vspace{1.5pt}
			\includegraphics[width=1.15\linewidth]{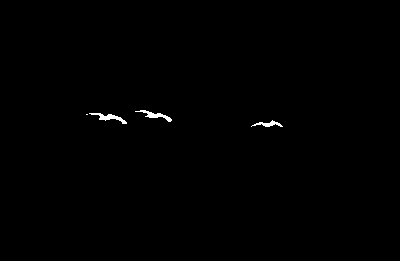}\vspace{1.5pt}
			\includegraphics[width=1.15\linewidth]{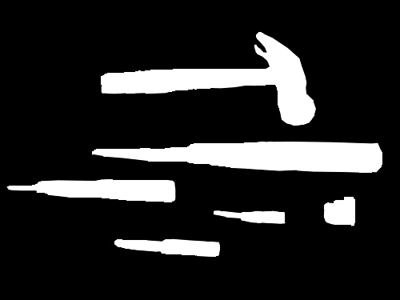}\vspace{1.5pt}
			\includegraphics[width=1.15\linewidth]{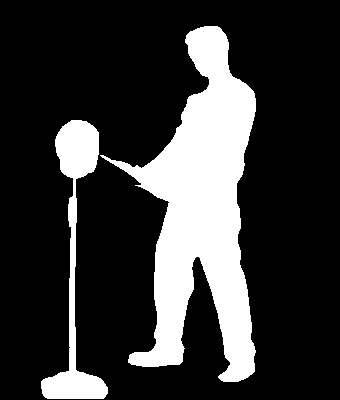}\vspace{1.5pt}
			\includegraphics[width=1.15\linewidth]{fig1/Mask/ILSVRC2012_test_00061303.png}\vspace{1.5pt}
			\includegraphics[width=1.15\linewidth]{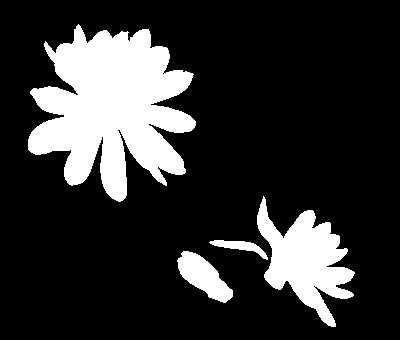}\vspace{1.5pt}
			\includegraphics[width=1.15\linewidth]{fig1/Mask/6681.png}\vspace{1.5pt}
			\includegraphics[width=1.15\linewidth]{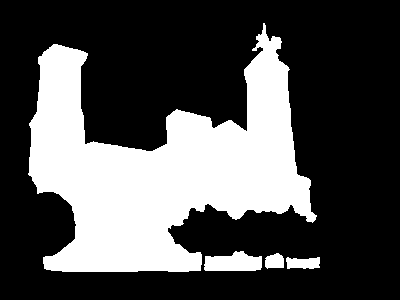}\vspace{1.5pt}
			\includegraphics[width=1.15\linewidth]{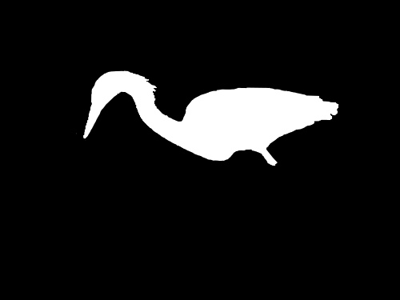}\vspace{1.5pt}
			\includegraphics[width=1.15\linewidth]{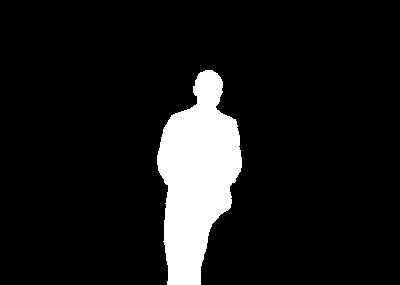}	
	\end{minipage}}
	\subfigure[\tiny Ours]{
		\begin{minipage}[b]{0.068\linewidth}
			\includegraphics[width=1.15\linewidth]{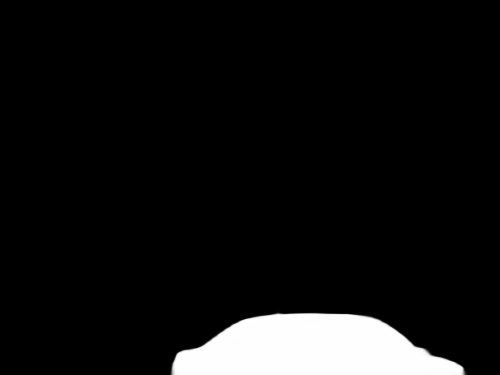}\vspace{1.5pt}
			\includegraphics[width=1.15\linewidth]{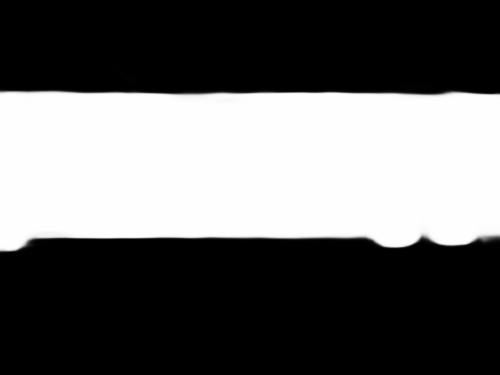}\vspace{1.5pt}
			\includegraphics[width=1.15\linewidth]{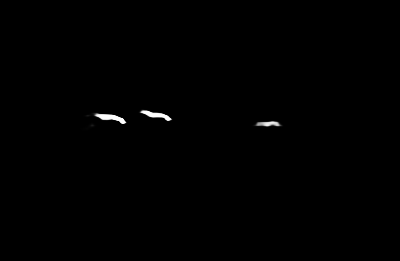}\vspace{1.5pt}
			\includegraphics[width=1.15\linewidth]{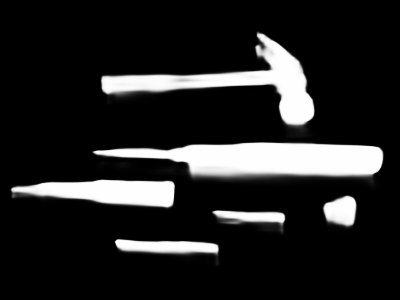}\vspace{1.5pt}
			\includegraphics[width=1.15\linewidth]{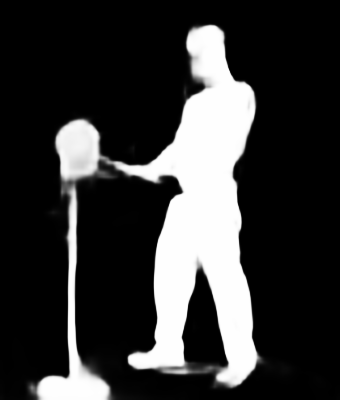}\vspace{1.5pt}
			\includegraphics[width=1.15\linewidth]{fig1/Ours/ILSVRC2012_test_00061303.png}\vspace{1.5pt}
			\includegraphics[width=1.15\linewidth]{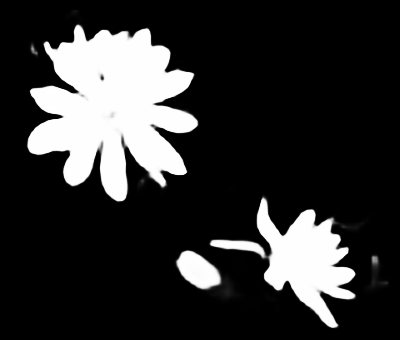}\vspace{1.5pt}
			\includegraphics[width=1.15\linewidth]{fig1/Ours/6681.png}\vspace{1.5pt}
			\includegraphics[width=1.15\linewidth]{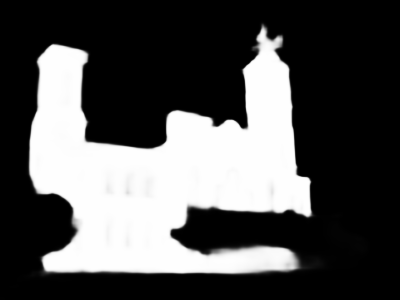}\vspace{1.5pt}
			\includegraphics[width=1.15\linewidth]{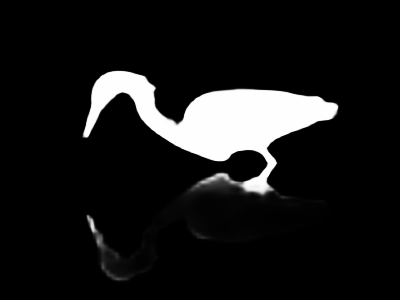}\vspace{1.5pt}
			\includegraphics[width=1.15\linewidth]{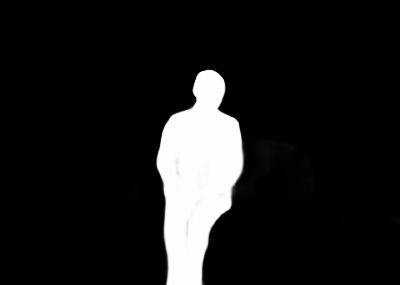}	
	\end{minipage}}
	\subfigure[\tiny ICON \cite{zhuge2021salient}]{
		\begin{minipage}[b]{0.068\linewidth}			
			\includegraphics[width=1.15\linewidth]{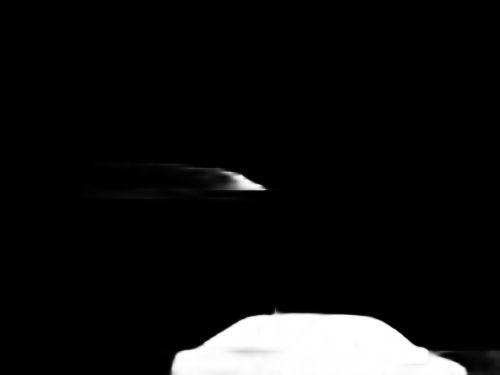}\vspace{1.5pt}
			\includegraphics[width=1.15\linewidth]{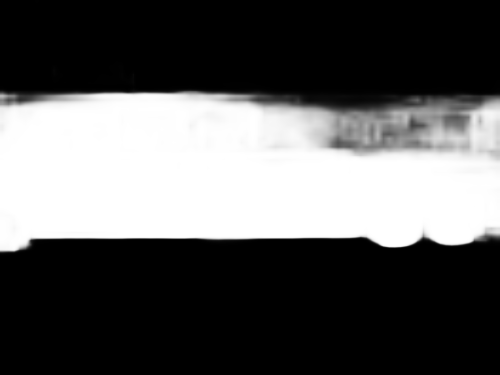}\vspace{1.5pt}
			\includegraphics[width=1.15\linewidth]{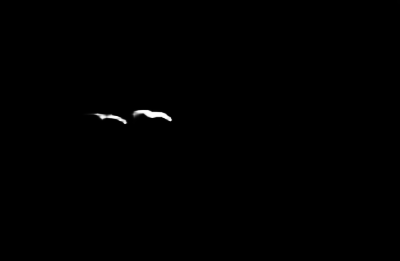}\vspace{1.5pt}
			\includegraphics[width=1.15\linewidth]{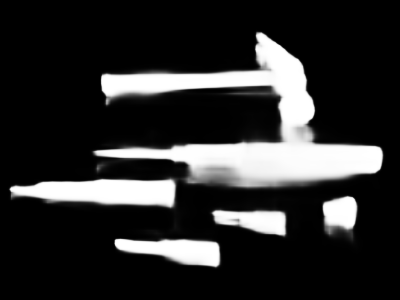}\vspace{1.5pt}
			\includegraphics[width=1.15\linewidth]{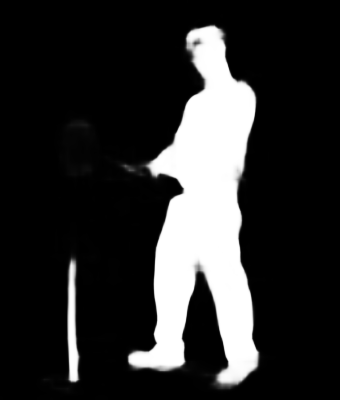}\vspace{1.5pt}
			\includegraphics[width=1.15\linewidth]{fig1/ICON/ILSVRC2012_test_00061303.png}\vspace{1.5pt}
			\includegraphics[width=1.15\linewidth]{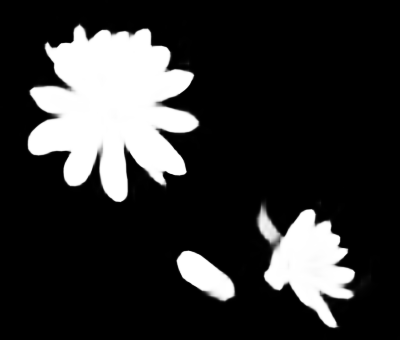}\vspace{1.5pt}
			\includegraphics[width=1.15\linewidth]{fig1/ICON/6681.png}\vspace{1.5pt}
			\includegraphics[width=1.15\linewidth]{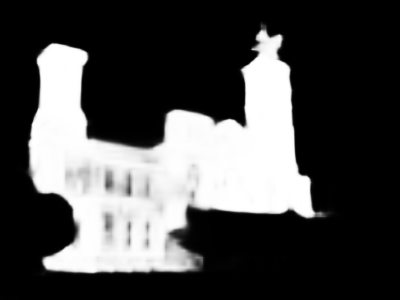}\vspace{1.5pt}
			\includegraphics[width=1.15\linewidth]{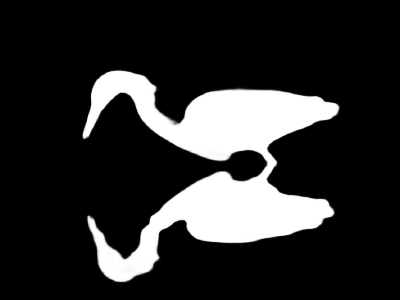}\vspace{1.5pt}
			\includegraphics[width=1.15\linewidth]{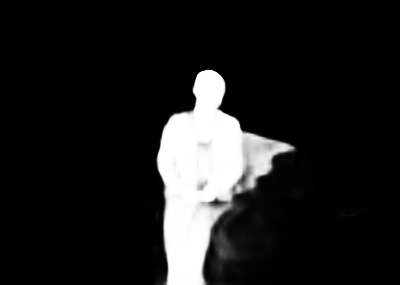}		
	\end{minipage}}	
	\subfigure[\tiny DCN \cite{DBLP:journals/tip/WuSH21}]{
		\begin{minipage}[b]{0.068\linewidth}			
			\includegraphics[width=1.15\linewidth]{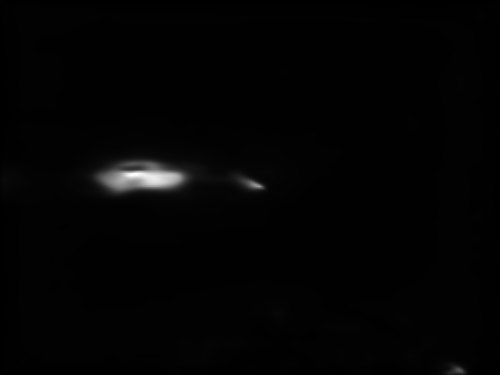}\vspace{1.5pt}
			\includegraphics[width=1.15\linewidth]{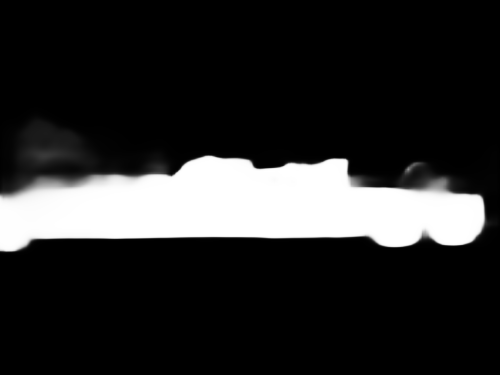}\vspace{1.5pt}
			\includegraphics[width=1.15\linewidth]{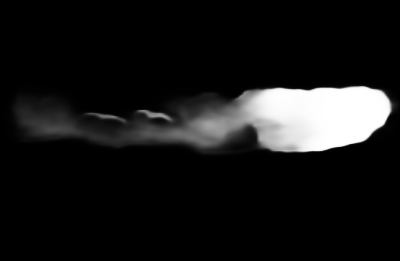}\vspace{1.5pt}
			\includegraphics[width=1.15\linewidth]{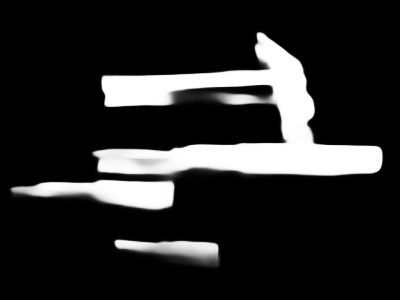}\vspace{1.5pt}
			\includegraphics[width=1.15\linewidth]{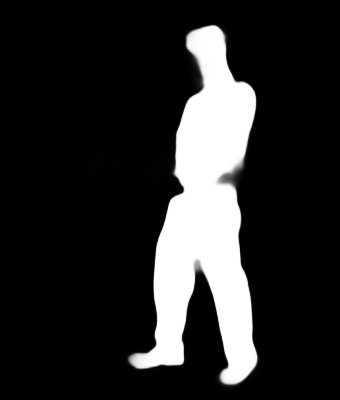}\vspace{1.5pt}
			\includegraphics[width=1.15\linewidth]{fig1/DCN/ILSVRC2012_test_00061303.png}\vspace{1.5pt}
			\includegraphics[width=1.15\linewidth]{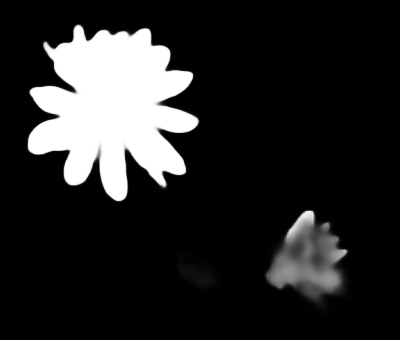}\vspace{1.5pt}
			\includegraphics[width=1.15\linewidth]{fig1/DCN/6681.png}\vspace{1.5pt}
			\includegraphics[width=1.15\linewidth]{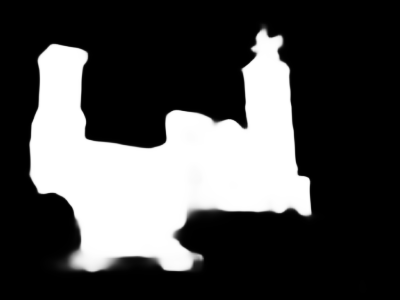}\vspace{1.5pt}
			\includegraphics[width=1.15\linewidth]{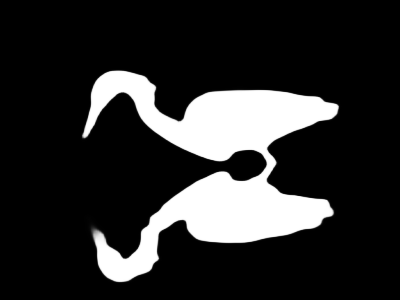}\vspace{1.5pt}
			\includegraphics[width=1.15\linewidth]{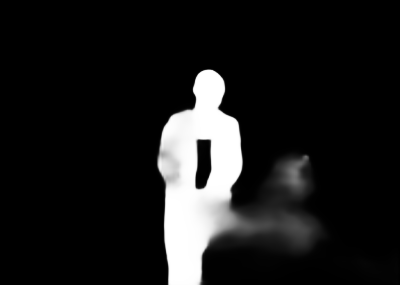}				
	\end{minipage}}
	\subfigure[\tiny MSFNet \cite{DBLP:conf/mm/ZhangLPYL21}]{
		\begin{minipage}[b]{0.068\linewidth}
			\includegraphics[width=1.15\linewidth]{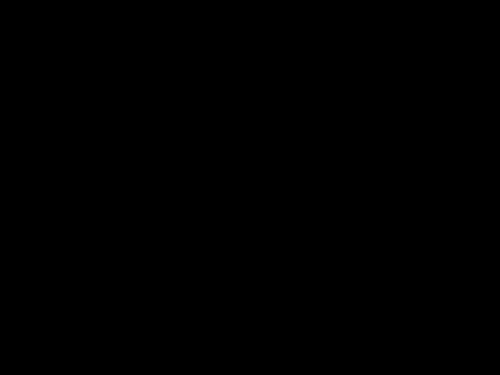}\vspace{1.5pt}
			\includegraphics[width=1.15\linewidth]{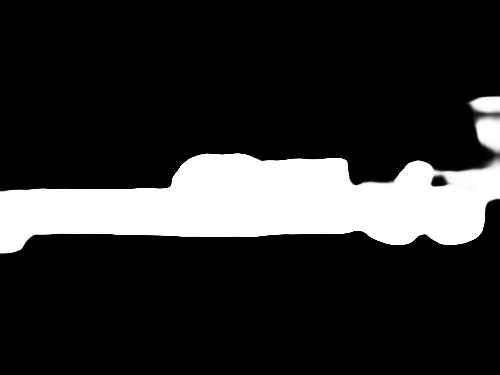}\vspace{1.5pt}
			\includegraphics[width=1.15\linewidth]{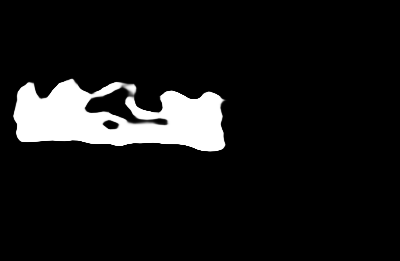}\vspace{1.5pt}
			\includegraphics[width=1.15\linewidth]{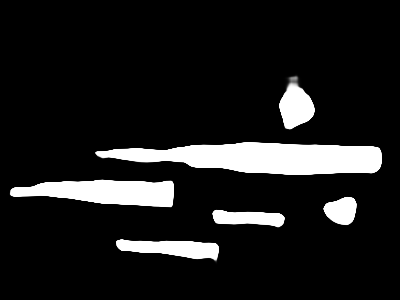}\vspace{1.5pt}
			\includegraphics[width=1.15\linewidth]{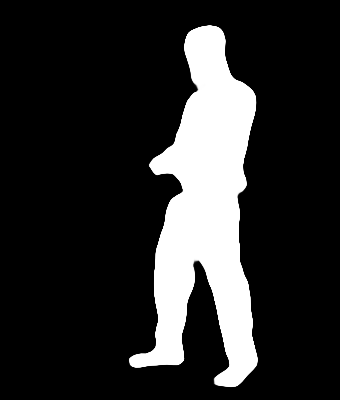}\vspace{1.5pt}
			\includegraphics[width=1.15\linewidth]{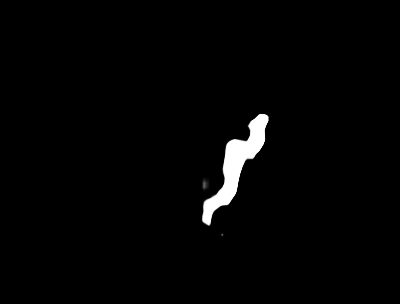}\vspace{1.5pt}
			\includegraphics[width=1.15\linewidth]{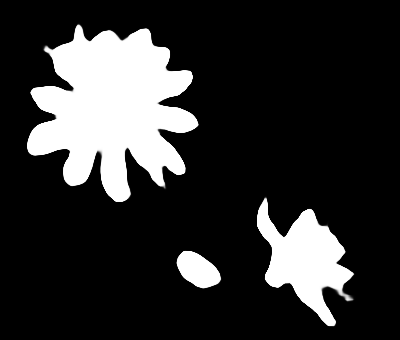}\vspace{1.5pt}
			\includegraphics[width=1.15\linewidth]{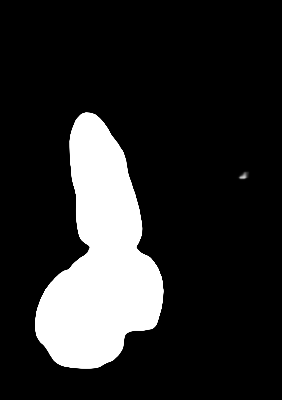}\vspace{1.5pt}
			\includegraphics[width=1.15\linewidth]{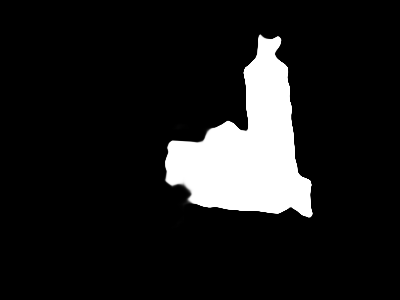}\vspace{1.5pt}
			\includegraphics[width=1.15\linewidth]{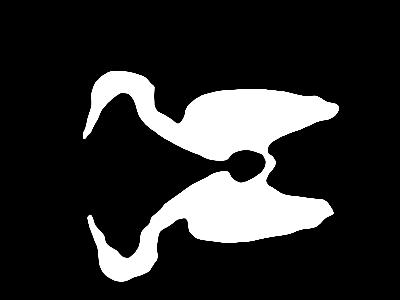}\vspace{1.5pt}
			\includegraphics[width=1.15\linewidth]{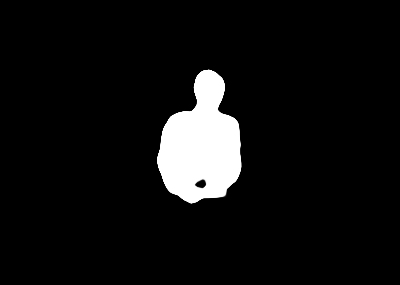}	
	\end{minipage}}
	\subfigure[\tiny PurNet \cite{DBLP:journals/tip/LiSXMT21}]{
		\begin{minipage}[b]{0.068\linewidth}			
			\includegraphics[width=1.15\linewidth]{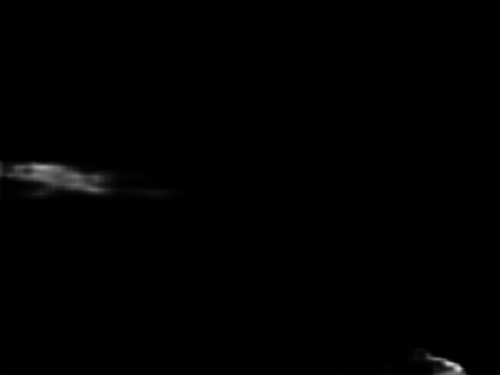}\vspace{1.5pt}
			\includegraphics[width=1.15\linewidth]{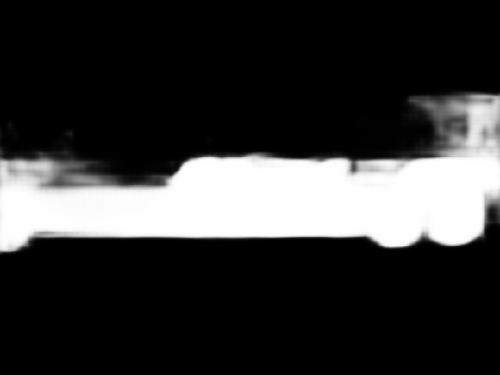}\vspace{1.5pt}
			\includegraphics[width=1.15\linewidth]{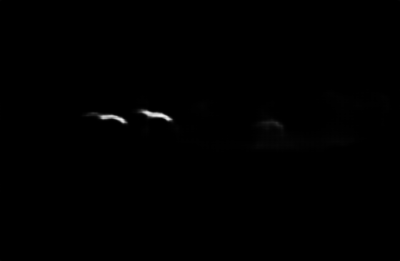}\vspace{1.5pt}
			\includegraphics[width=1.15\linewidth]{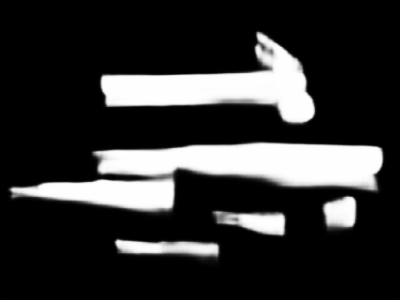}\vspace{1.5pt}
			\includegraphics[width=1.15\linewidth]{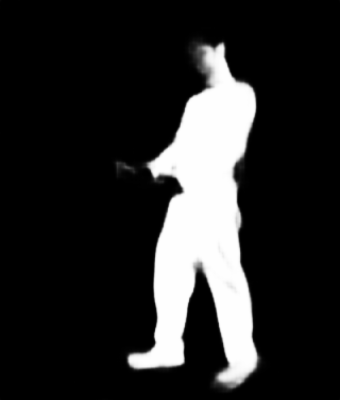}\vspace{1.5pt}
			\includegraphics[width=1.15\linewidth]{fig1/PurNet/ILSVRC2012_test_00061303.png}\vspace{1.5pt}
			\includegraphics[width=1.15\linewidth]{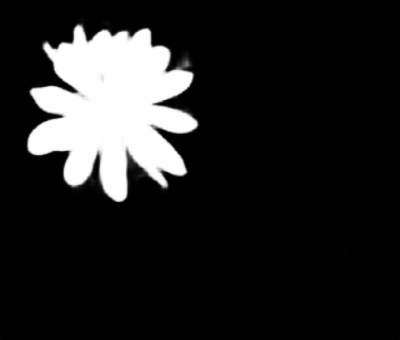}\vspace{1.5pt}
			\includegraphics[width=1.15\linewidth]{fig1/PurNet/6681.png}\vspace{1.5pt}
			\includegraphics[width=1.15\linewidth]{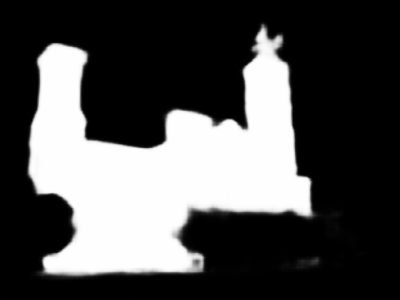}\vspace{1.5pt}
			\includegraphics[width=1.15\linewidth]{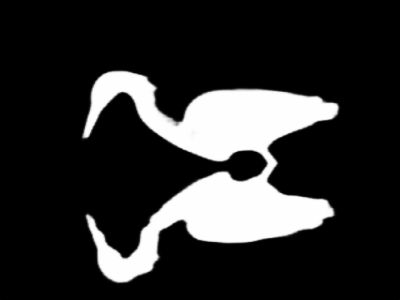}\vspace{1.5pt}
			\includegraphics[width=1.15\linewidth]{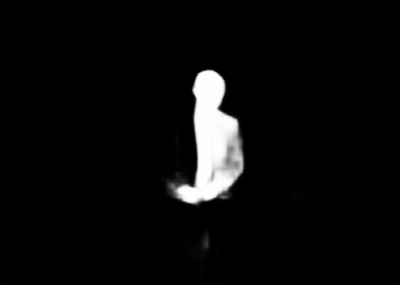}				
	\end{minipage}}
	\subfigure[\tiny GFINet \cite{zhu2021supplement}]{
		\begin{minipage}[b]{0.068\linewidth}			
			\includegraphics[width=1.15\linewidth]{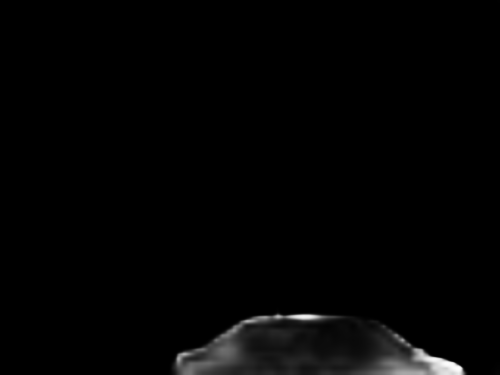}\vspace{1.5pt}
			\includegraphics[width=1.15\linewidth]{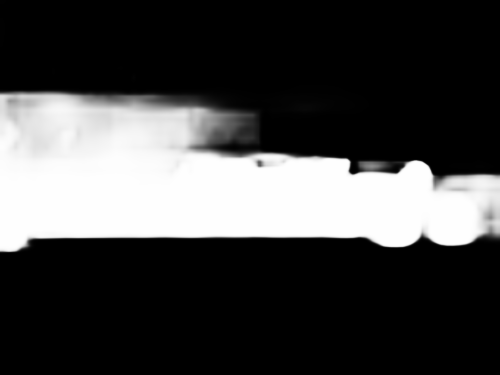}\vspace{1.5pt}
			\includegraphics[width=1.15\linewidth]{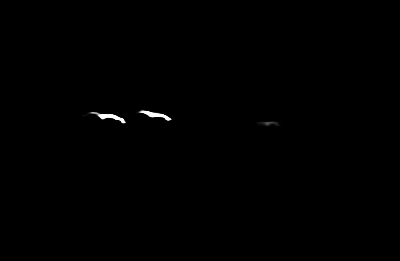}\vspace{1.5pt}
			\includegraphics[width=1.15\linewidth]{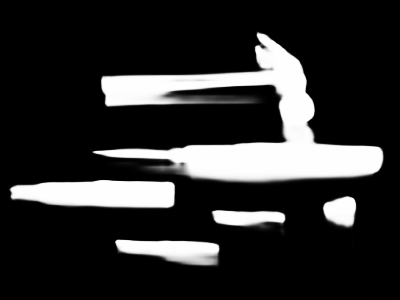}\vspace{1.5pt}
			\includegraphics[width=1.15\linewidth]{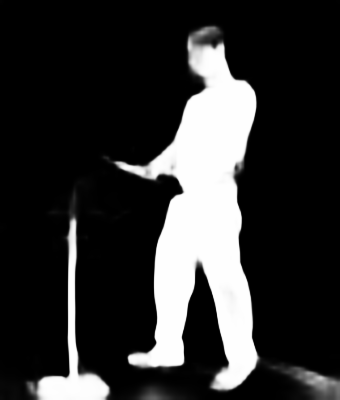}\vspace{1.5pt}
			\includegraphics[width=1.15\linewidth]{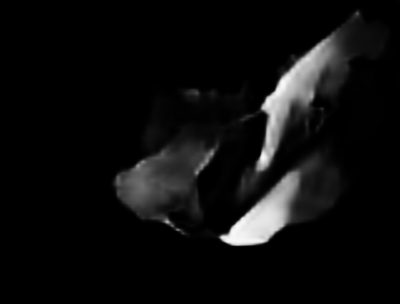}\vspace{1.5pt}
			\includegraphics[width=1.15\linewidth]{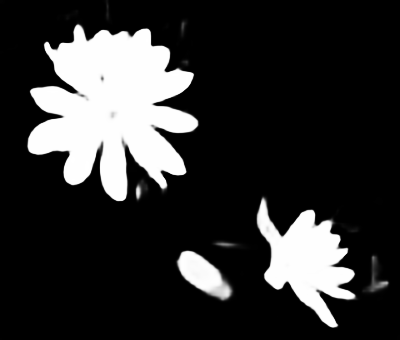}\vspace{1.5pt}
			\includegraphics[width=1.15\linewidth]{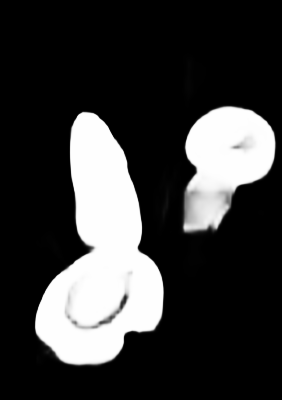}\vspace{1.5pt}
			\includegraphics[width=1.15\linewidth]{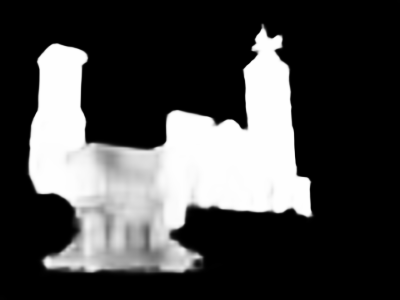}\vspace{1.5pt}
			\includegraphics[width=1.15\linewidth]{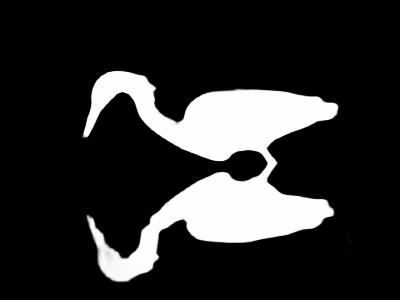}\vspace{1.5pt}
			\includegraphics[width=1.15\linewidth]{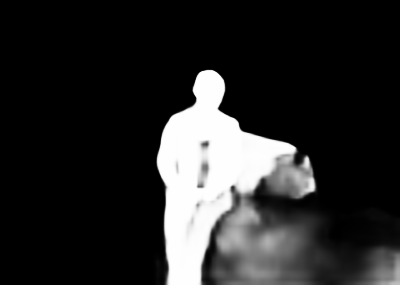}			
	\end{minipage}}
	\subfigure[\tiny ITSD \cite{DBLP:conf/cvpr/ZhouXLCY20}]{
		\begin{minipage}[b]{0.068\linewidth}
			
			\includegraphics[width=1.15\linewidth]{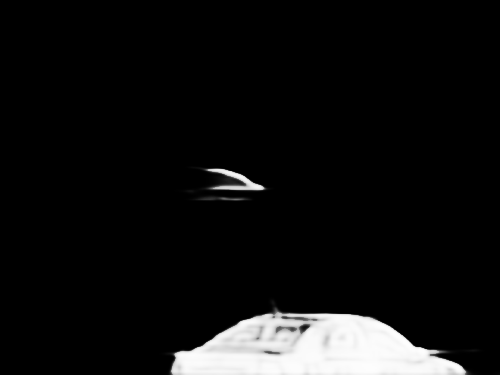}\vspace{1.5pt}
			\includegraphics[width=1.15\linewidth]{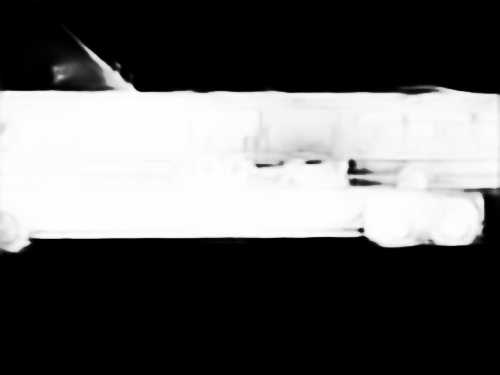}\vspace{1.5pt}
			\includegraphics[width=1.15\linewidth]{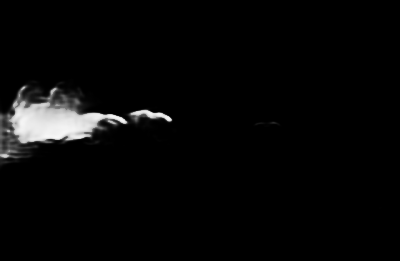}\vspace{1.5pt}
			\includegraphics[width=1.15\linewidth]{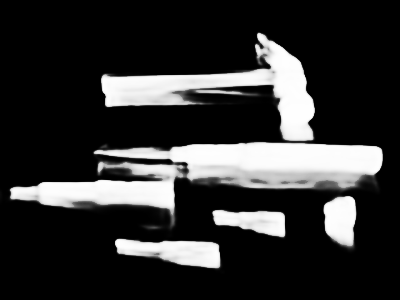}\vspace{1.5pt}
			\includegraphics[width=1.15\linewidth]{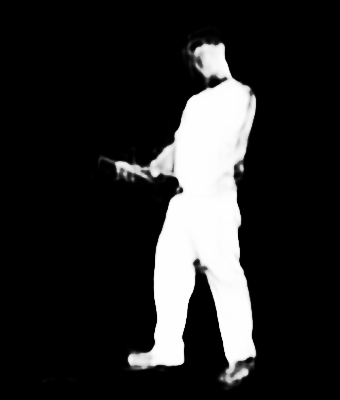}\vspace{1.5pt}
			\includegraphics[width=1.15\linewidth]{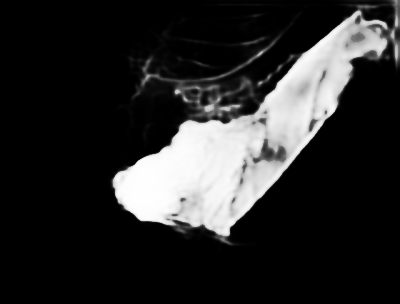}\vspace{1.5pt}
			\includegraphics[width=1.15\linewidth]{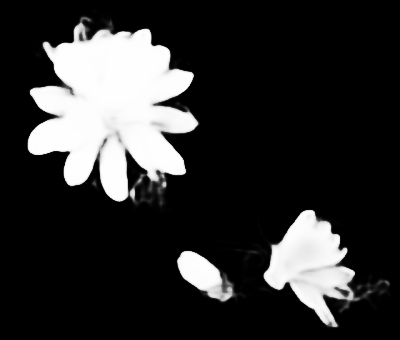}\vspace{1.5pt}
			\includegraphics[width=1.15\linewidth]{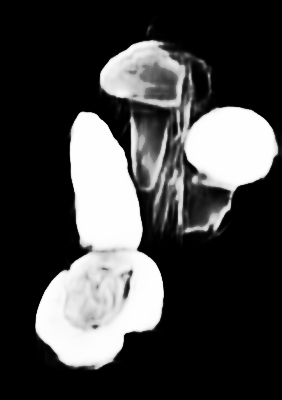}\vspace{1.5pt}
			\includegraphics[width=1.15\linewidth]{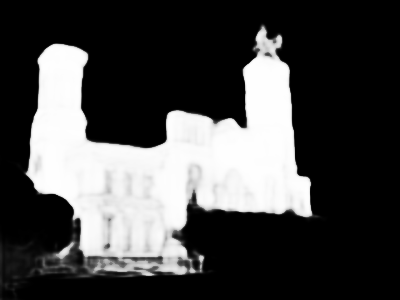}\vspace{1.5pt}
			\includegraphics[width=1.15\linewidth]{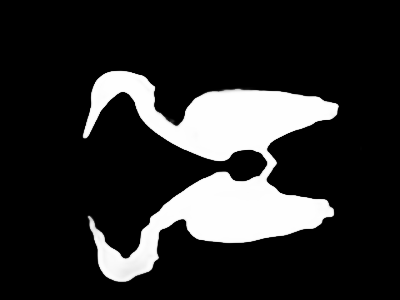}\vspace{1.5pt}
			\includegraphics[width=1.15\linewidth]{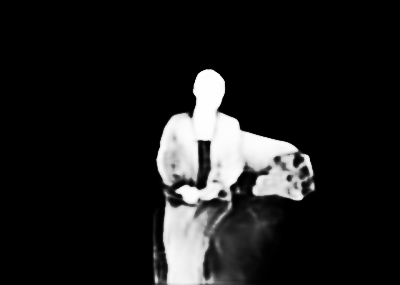}
	\end{minipage}}
	\subfigure[\tiny MINet \cite{DBLP:conf/cvpr/PangZZL20}]{
		\begin{minipage}[b]{0.068\linewidth}
			\includegraphics[width=1.15\linewidth]{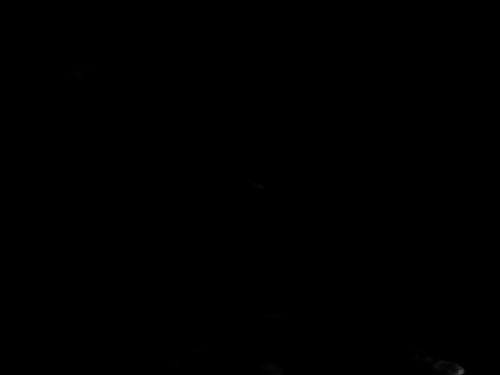}\vspace{1.5pt}
			\includegraphics[width=1.15\linewidth]{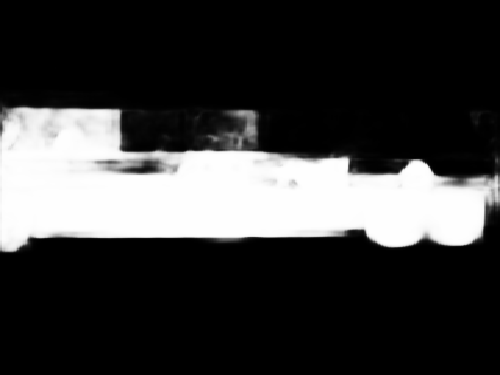}\vspace{1.5pt}
			\includegraphics[width=1.15\linewidth]{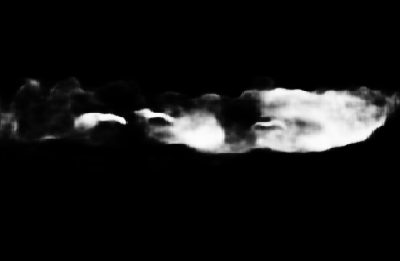}\vspace{1.5pt}
			\includegraphics[width=1.15\linewidth]{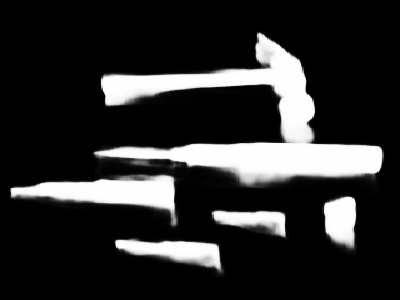}\vspace{1.5pt}
			\includegraphics[width=1.15\linewidth]{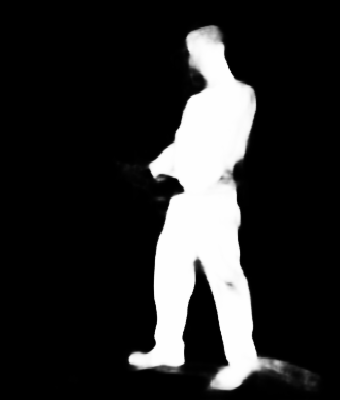}\vspace{1.5pt}
			\includegraphics[width=1.15\linewidth]{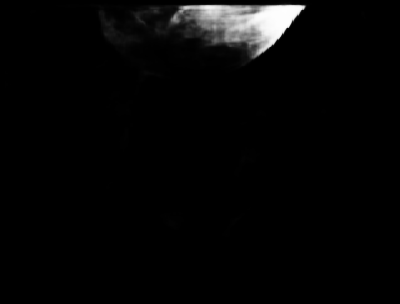}\vspace{1.5pt}
			\includegraphics[width=1.15\linewidth]{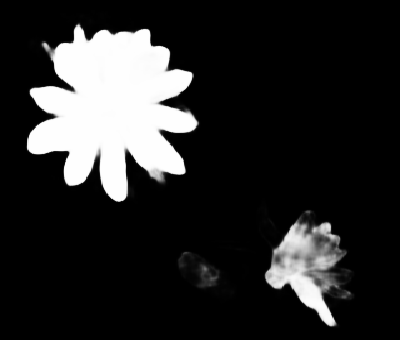}\vspace{1.5pt}
			\includegraphics[width=1.15\linewidth]{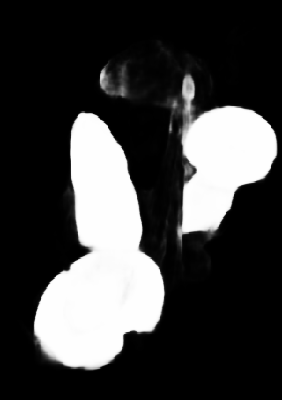}\vspace{1.5pt}
			\includegraphics[width=1.15\linewidth]{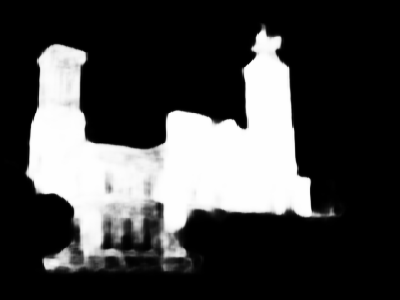}\vspace{1.5pt}
			\includegraphics[width=1.15\linewidth]{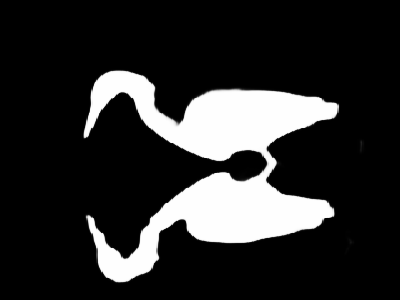}\vspace{1.5pt}
			\includegraphics[width=1.15\linewidth]{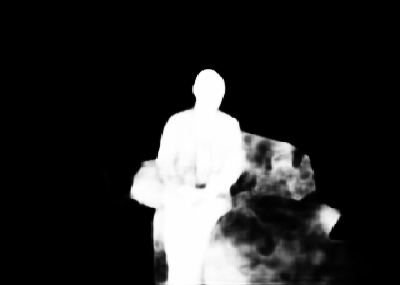}
	\end{minipage}}
	\subfigure[\tiny GateNet \cite{DBLP:conf/eccv/ZhaoPZLZ20}]{
		\begin{minipage}[b]{0.068\linewidth}			
			\includegraphics[width=1.15\linewidth]{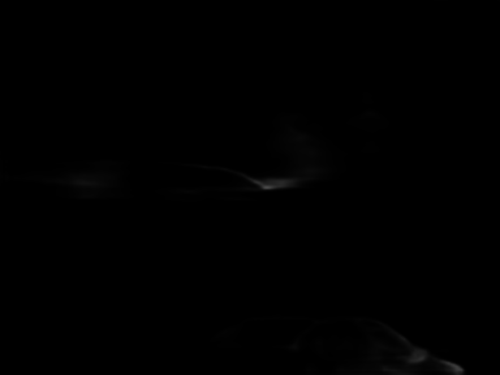}\vspace{1.5pt}
			\includegraphics[width=1.15\linewidth]{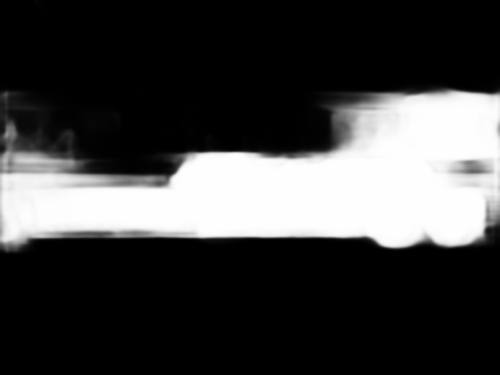}\vspace{1.5pt}
			\includegraphics[width=1.15\linewidth]{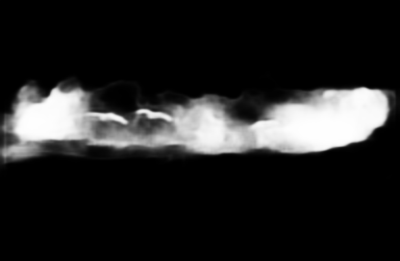}\vspace{1.5pt}
			\includegraphics[width=1.15\linewidth]{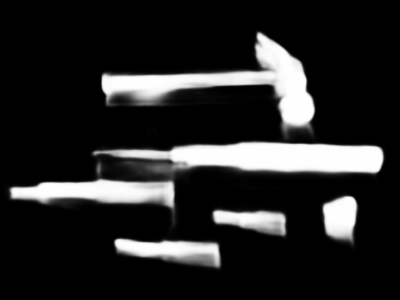}\vspace{1.5pt}
			\includegraphics[width=1.15\linewidth]{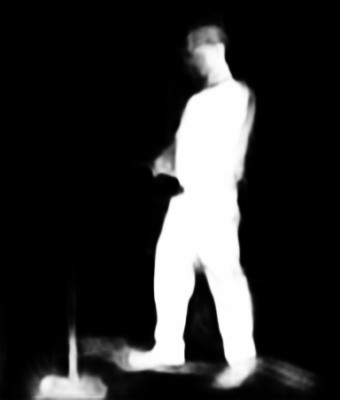}\vspace{1.5pt}
			\includegraphics[width=1.15\linewidth]{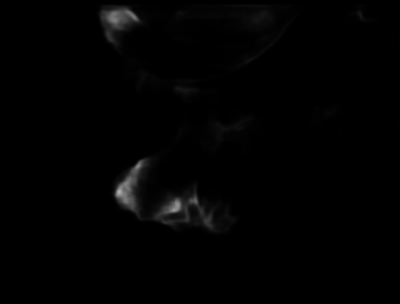}\vspace{1.5pt}
			\includegraphics[width=1.15\linewidth]{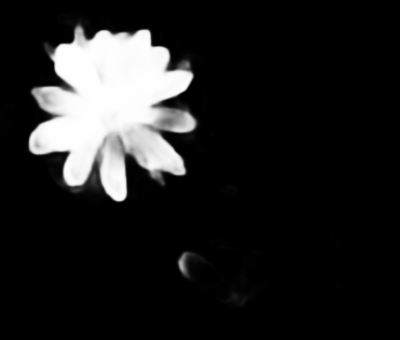}\vspace{1.5pt}
			\includegraphics[width=1.15\linewidth]{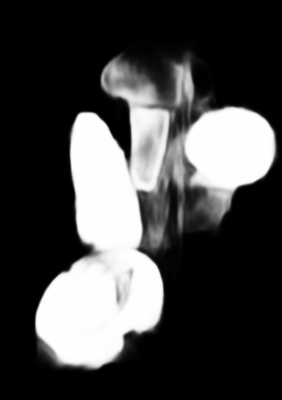}\vspace{1.5pt}
			\includegraphics[width=1.15\linewidth]{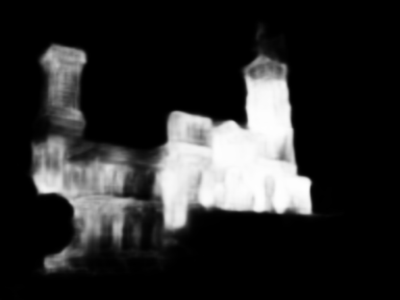}\vspace{1.5pt}
			\includegraphics[width=1.15\linewidth]{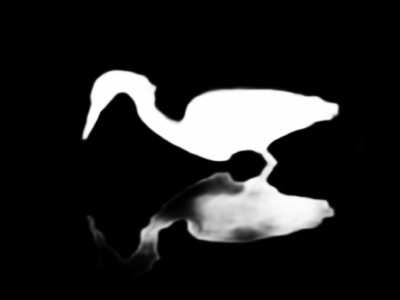}\vspace{1.5pt}
			\includegraphics[width=1.15\linewidth]{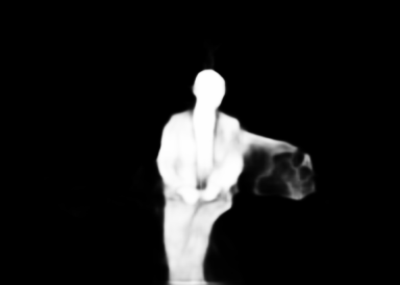}
	\end{minipage}}
	\subfigure[\tiny F3Net \cite{DBLP:conf/aaai/WeiWH20}]{
		\begin{minipage}[b]{0.068\linewidth}
			\includegraphics[width=1.15\linewidth]{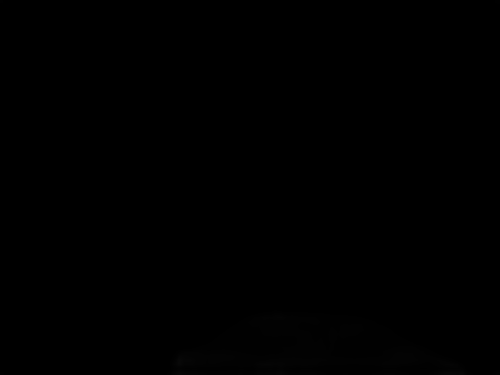}\vspace{1.5pt}
			\includegraphics[width=1.15\linewidth]{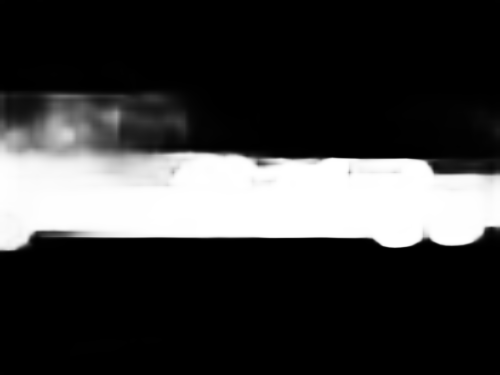}\vspace{1.5pt}
			\includegraphics[width=1.15\linewidth]{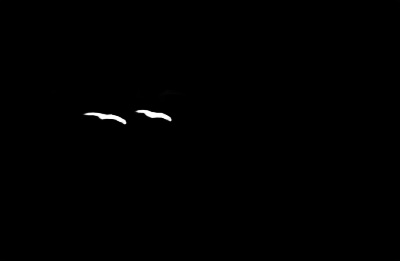}\vspace{1.5pt}
			\includegraphics[width=1.15\linewidth]{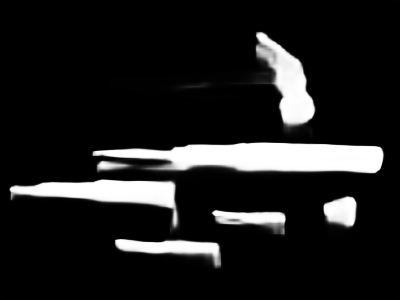}\vspace{1.5pt}
			\includegraphics[width=1.15\linewidth]{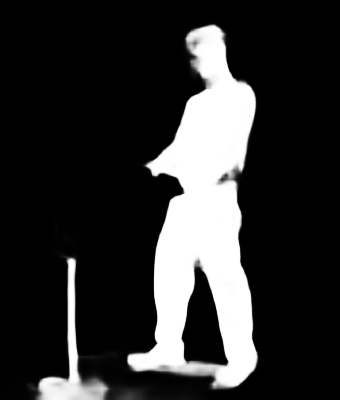}\vspace{1.5pt}
			\includegraphics[width=1.15\linewidth]{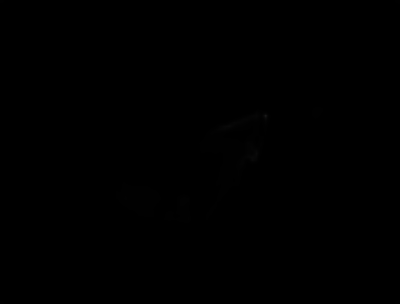}\vspace{1.5pt}
			\includegraphics[width=1.15\linewidth]{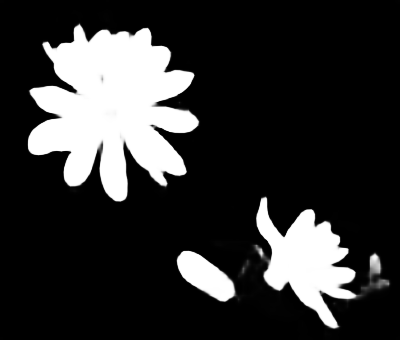}\vspace{1.5pt}
			\includegraphics[width=1.15\linewidth]{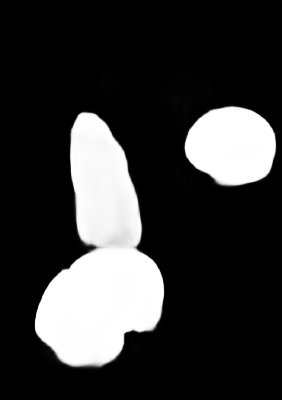}\vspace{1.5pt}
			\includegraphics[width=1.15\linewidth]{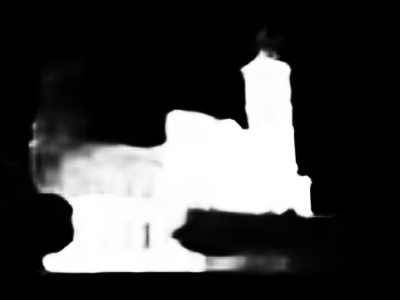}\vspace{1.5pt}
			\includegraphics[width=1.15\linewidth]{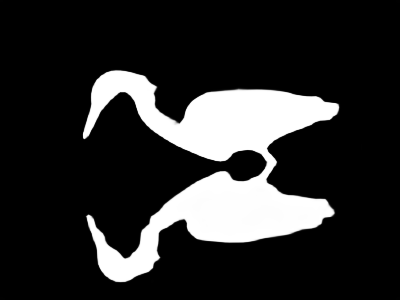}\vspace{1.5pt}
			\includegraphics[width=1.15\linewidth]{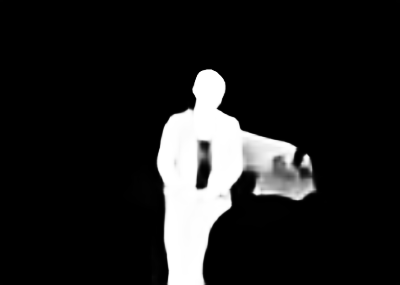}
	\end{minipage}}	
	\caption{Qualitative comparisons with state-of-the-arts in different complex scenes. From top to bottom are out-of-view, large object, small object, multi-object, thin object, low contrast, shape complexity, cluttered background, occlusion, reflection and foreground disturbance, respectively.}
\end{figure*}
\section{Experiments}
In this section, we first introduce the used datasets and evaluation metrics. Then, we describe the implementation details. Finally, we present extensive experimental results to demonstrate the effectiveness of the proposed method.
\subsection{Datasets}
To evaluate the performance of our model, we conduct experiments on five benchmark datasets. Among them, ECSSD \cite{DBLP:conf/cvpr/YanXSJ13} contains 1000 images with semantically meaningful images but structurally complex objects. HKU-IS \cite{DBLP:journals/tip/LiY16} is composed of 4,447 complex scenes, which contains multiple disconnected objects and has a similar appearance between foreground and background. PASCAL-S \cite{DBLP:conf/cvpr/LiHKRY14}  with 850 challenging images that are carefully selected from the validation set of PASCAL VOC \cite{DBLP:journals/ijcv/EveringhamGWWZ10}. DUT-OMRON \cite{DBLP:conf/cvpr/YangZLRY13} consists of 5,168 high-quality images which have relatively cluttered background and high content variety. DUTS \cite{DBLP:conf/cvpr/WangLWF0YR17} is currently the largest SOD dataset, which contains 10,553 training images (DUTS-TR) and 5,019 test images (DUTS-TE). Both training and test sets contain complex scenes. We use the DUTS-TR as the training dataset and others as the test datasets. 
\subsection{Evaluation Metrics}
To evaluate the performance of our method, we adopt five widespread metrics for a comprehensive comparison.\\
\textbf{Precision-Recall (PR)} curve is calculated based on binary saliency map and ground truth:
\begin{equation}
{\mathop{\rm Precision}\nolimits}  = {\rm{ }}\frac{{{\mathop{\rm TP}\nolimits} }}{{{\mathop{\rm TP}\nolimits}  + {\mathop{\rm FP}\nolimits} }},{\rm{ Recall  =  }}\frac{{{\mathop{\rm TP}\nolimits} }}{{{\mathop{\rm TP}\nolimits}  + {\mathop{\rm FN}\nolimits} }},
\end{equation}
where TP, TN, FP, FN denote true positive, true negative, false positive and false negative, respectively. A set of thresholds ranging from 0 to 255 is used to binarize the prediction. For each threshold, we can get a pair of precision/recall and then form a PR curve.\\
\textbf{Mean absolute error (MAE)} measures the average L1 distance between prediction and ground truth:
\begin{equation}
{\mathop{\rm MAE}\nolimits}  = \frac{1}{{H \cdot W}}\sum\limits_{i = 1}^H {\sum\limits_{j = 1}^W {|P(i,j) - GT(i,j)} } |,
\end{equation}
where $P$ is the predicted saliency map and $GT$ is the ground truth.\\
\textbf{F-measure} considers both precision and recall by computing the weighted harmonic mean:
\begin{equation}
{F_\beta } = \frac{{{{(1 + \beta )}^2} \cdot Precision \cdot Recall}}{{{\beta ^2} \cdot Precision + Recall}},
\end{equation}
where $\beta$ is set to 0.3 to emphasize the precision over recall. We plot the whole F-measure curve and report the maximum F-measure score ($F_{\beta}$), respectively.\\
\textbf{Structure similarity ($S_{\alpha}$)} \cite{DBLP:conf/iccv/FanCLLB17} evaluates structural similarity between predicted saliency map and ground-truth. $S_{\alpha}$ considers object-aware (${S_o}$) and region-aware (${S_r}$) structure similarities, respectively:
\begin{equation}
{S_m} = \alpha  \cdot {S_o} + (1 - \alpha ) \cdot {S_r},
\end{equation}
where $\alpha$ is set to 0.5 to balance the ${S_o}$ and ${S_r}$.
\subsection{Implementation Details}
Pytorch 1.3 is used to implement the proposed model. The backbone parameters are initialized from the ResNet-50 pre-trained on the ImageNet \cite{DBLP:conf/cvpr/DengDSLL009}, and other parameters are randomly initialized. During training, we resize each image to $320 \times 320$ with random horizontal flipping and hidden patch \cite{DBLP:journals/corr/abs-1811-02545}, and then randomly crop a patch with the size of $288 \times 288$. We set the maximum learning rate to 5e-3 for backbone and 5e-2 for other parts. Warm-up and linear decay strategies are used to adjust the learning rate. The whole network is trained end-to-end by stochastic gradient descent. Momentum and weight decay are set to 0.9 and 5e-4, respectively. Batch size is set to 64 and maximum epoch is set to 36. During testing, each image is simply resized to $320 \times 320$ and then fed into the network to predict. We don’t use any multi-scale training or any post-processing procedures (e.g. CRF \cite{DBLP:conf/nips/KrahenbuhlK11}). 
\subsection{Comparison with State-of-the-arts}
We compare the proposed method with 15 state-of-the-art SOD methods based on the ResNet backbone \cite{DBLP:conf/cvpr/HeZRS16}, including CPD-R \cite{DBLP:conf/cvpr/WuSH19}, BASNet \cite{DBLP:conf/cvpr/QinZHGDJ19}, PoolNet \cite{DBLP:conf/cvpr/LiuHCFJ19}, SIBI \cite{DBLP:conf/iccv/SuLZXT19}, CAGNet \cite{DBLP:journals/pr/MohammadiNBMH20}, ITSD \cite{DBLP:conf/cvpr/ZhouXLCY20}, MINet \cite{DBLP:conf/cvpr/PangZZL20}, GateNet \cite{DBLP:conf/eccv/ZhaoPZLZ20}, GCPA \cite{DBLP:conf/aaai/ChenXCH20}, F3Net \cite{DBLP:conf/aaai/WeiWH20}, PurNet \cite{DBLP:journals/tip/LiSXMT21}, GFINet \cite{zhu2021supplement}, MSFNet \cite{DBLP:conf/mm/ZhangLPYL21}, DCN \cite{DBLP:journals/tip/WuSH21} and ICON \cite{zhuge2021salient}. For fair comparison, we evaluate the best saliency maps provided by the authors with the same code.
\subsubsection{Quantitative Comparison}
Tab. \uppercase\expandafter{\romannumeral1} shows the quantitative comparison results in terms of three evaluation metrics on five benchmark datasets. It can be seen that our method consistently outperforms other competitors across three metrics on most datasets. Compared with the results of other methods that achieve an advantage in a certain metric on a certain dataset (e.g. MSFNet \cite{DBLP:conf/mm/ZhangLPYL21} in terms of MAE), our method has better generalization, which ranks first in most cases and second in a few cases across all three metrics on five datasets. Since $F_{\beta}$ is sensitive to various flaws of saliency maps, $S_m$ evaluates the structural integrity of the predicted saliency maps and MAE can indicate the strength of background noise, these three metrics comprehensively reflect whether a method can have a relatively complete prediction on the premise of low background noises. Compared with the ICON \cite{zhuge2021salient} which based on the method of aggregating multi-level features, GFINet \cite{zhu2021supplement} which introduces edge information and the DCN \cite{DBLP:journals/tip/WuSH21} which introduces both edge and skeleton information, the performance of our method are averagely improved by $2.1\%$, $1.9\%$ and $0.3\%$ respectively in terms of $F_{\beta}$, $S_m$ and MAE on five datasets. Especially, on the larger datasets (DUT-OMRON \cite{DBLP:conf/cvpr/YangZLRY13} and DUTS-TE \cite{DBLP:conf/cvpr/WangLWF0YR17}) which contain cluttered background and multiple salient objects with complex structure, our method has more obvious advantages than the above three methods, and the performance of our model are averagely improved by $3.9\%$, $3.2\%$ and $0.7\%$ respectively in terms of $F_{\beta}$, $S_m$ and MAE. The consistent improvement in above three metrics demonstrate that our method can effectively improve the completeness of prediction under the premise of low background noise.
\par Furthermore, we plot the precision-recall and F-measures curves of above methods for overall comparisons, as shown in Fig. 6. From these curves, we can observe that our method outperforms other competitors under different thresholds in most cases, which is consistent with the measures reported in Tab. \uppercase\expandafter{\romannumeral1}.
\par In addition, we compare the inference speed of our method and boundary accuracy with state-of-the-art models on DUT-OMRON \cite{DBLP:conf/cvpr/YangZLRY13} dataset. $\rm MAE_{b}$ \cite{zhu2021supplement} is used to evaluate the quality of boundary, which represents the MAE of the region within 10 pixels from the boundary, including both inside and outside of the boundary. All models utilize ResNet-50 as encoder and their decoder consist of multiple branches, including the branch for predicting salient objects and the branches for predicting boundaries or skeletons. For fair comparisons, we test these models with the same code and environment. As presented in Tab. \uppercase\expandafter{\romannumeral2}, with $320 \times 320$ input images, our SENet runs at speed of 39.5 FPS on a single 2080Ti GPU and Intel Xeon CPU E5-2678 v3 @ 2.50GHz. Compared with the method DCN \cite{DBLP:journals/tip/WuSH21} based on introducing both edge and skeleton information, our SENet achieves superior performance over it on FPS, $\rm MAE_{b}$ and $F_{\beta}$.  Compared with the method GFINet \cite{zhu2021supplement} based on introducing edge information, our model achieves competitive result on $\rm MAE_{b}$. Although the speed of our method is a little slower than GFINet \cite{zhu2021supplement}, the completeness of prediction is better than it. In the process of decoding, our model gradually feed the expanded saliency features generated by TS branch into OS branch, which is also in line with the human visual system. Compared with other counterparts, our model not only achieves better boundary and overall accuracy but also is efficient.
\begin{table}[ht]
	\centering
	\setlength{\abovecaptionskip}{0pt}
	\setlength{\belowcaptionskip}{3pt}
	\renewcommand\arraystretch{1.15}
	\setlength\tabcolsep{5.6pt}
	\caption{Comparisons of inference speed and accuracy of different SOD methods on DUT-OMRON \cite{DBLP:conf/cvpr/YangZLRY13} dataset. All methods adopt ResNet-50 as encoder and their decoder consists of multiple branches for predicting. The best three results are marked in \textcolor{red}{red} and \textcolor{green}{green}.}
	\begin{tabular}{lccccc}
		\toprule
		Method & Encoder & Input Size   & FPS & $\rm MAE_{b}$ & $F_{\beta}$\\
		\midrule
		EGNet$_{19}$ \cite{DBLP:conf/iccv/ZhaoLFCYC19}   &ResNet-50& $380 \times 320$&	12.4 &	0.022 & 0.815\\	
		SIBI$_{19}$ \cite{DBLP:conf/iccv/SuLZXT19}   &ResNet-50& $400 \times 300$&	15.1& 0.023 & 0.803 \\		
		SCRN$_{19}$ \cite{DBLP:conf/iccv/WuSH19}   &ResNet-50&  $352 \times 352$& 37.1 & 0.024 & 0.812 \\
		ITSD$_{20}$ \cite{DBLP:conf/cvpr/ZhouXLCY20}   &ResNet-50&  $288 \times 288$	& 38.4 & 0.024&	0.821 \\
		DCN$_{21}$ \cite{DBLP:journals/tip/WuSH21}   &ResNet-50& $352 \times 352$& 31.0 & 0.020 & \textcolor{green}{0.823} \\	
		GFINet$_{21}$ \cite{zhu2021supplement}   &ResNet-50 & $320 \times 320$ & \textcolor{red}{46.5} 	&\textcolor{red}{0.019}&	\textcolor{green}{0.823} \\
		\textbf{SENet (Ours)}   &ResNet-50 & $320 \times 320$ & \textcolor{green}{39.5} &	\textcolor{red}{0.019}&	\textcolor{red}{0.827} \\		
		\bottomrule
	\end{tabular}
\end{table}
\subsubsection{Qualitative Comparison}
To further illustrate the advantages of our model, we provide the visual comparisons on 11 complex scenes, as shown in Fig. 7. From top to bottom are out-of-view, large object, small object, multi-object, thin object, low contrast, shape complexity, cluttered background, occlusion, reflection and foreground interference, respectively. These scenes have common characteristics, including cluttered background, low contrast, similar colors or textures between foreground and background. One can observe that the proposed method works better on detecting and segmenting salient objects, which not only obtains complete saliency maps, but also well suppresses background noises, such as the car at image boundary (row 1), the bus on the road (row 2), the birds in the clouds (row 3), the holder on the showroom (row 5), the frog on the leaves (row 6), the flowers on branches (row 7), the mushrooms in the bushes (row 8) and the building around the lawn (row 9). It illustrates the power of the proposed strategy of separation first and then segmentation, which can utilize the expanded saliency features to guide the network obtain complete saliency maps with low noises. 
\begin{figure*}[ht]
	\makeatletter
	\renewcommand{\@thesubfigure}{\hskip\subfiglabelskip}
	\makeatother		
	\subfigure[Image]{
		\begin{minipage}[b]{0.112\linewidth}					
			\includegraphics[width=1.25\linewidth]{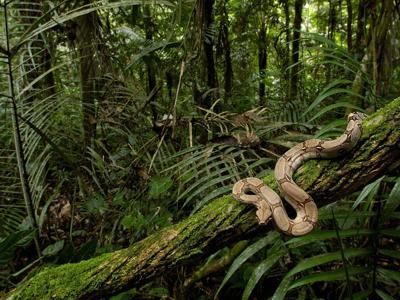}\vspace{1pt}		
			\includegraphics[width=1.25\linewidth]{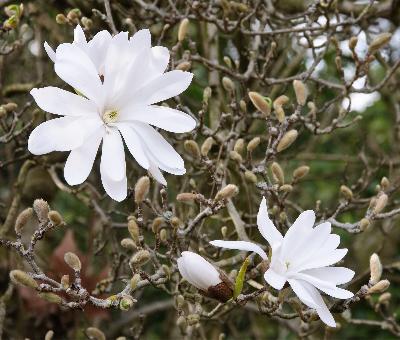}\vspace{1pt}
			\includegraphics[width=1.25\linewidth]{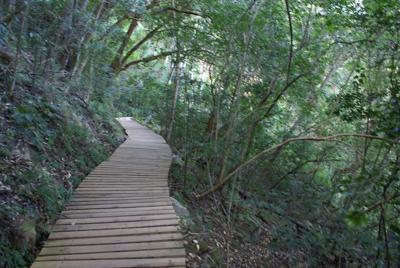}\vspace{1pt}
			\includegraphics[width=1.25\linewidth]{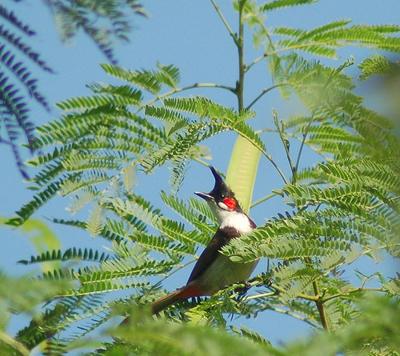}\vspace{1pt}	
	\end{minipage}}	
	\hspace{.07in}
	\subfigure[GT]{
		\begin{minipage}[b]{0.112\linewidth}			
			\includegraphics[width=1.25\linewidth]{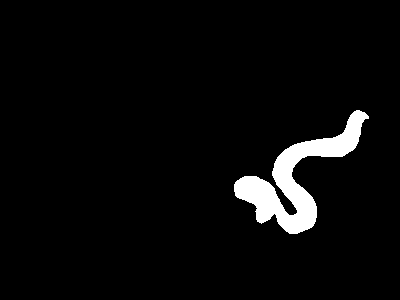}\vspace{1pt}		
			\includegraphics[width=1.25\linewidth]{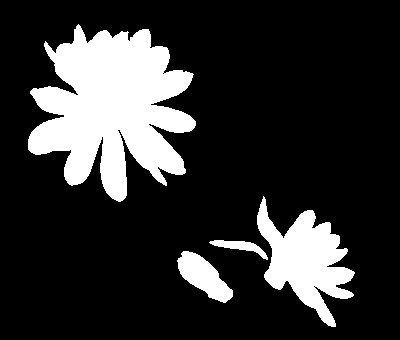}\vspace{1pt}
			\includegraphics[width=1.25\linewidth]{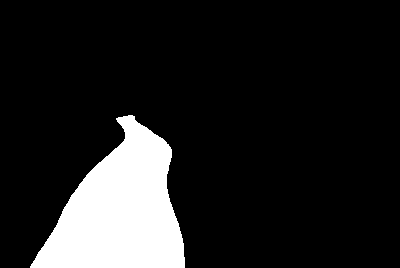}\vspace{1pt}				
			\includegraphics[width=1.25\linewidth]{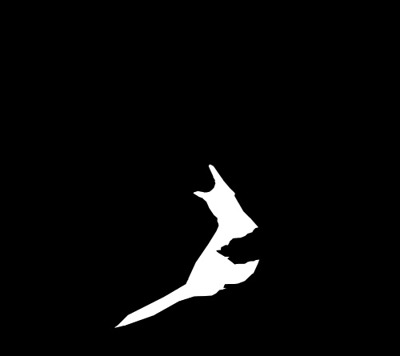}\vspace{1pt}
	\end{minipage}}	
	\hspace{.07in}
	\subfigure[Baseline]{
		\begin{minipage}[b]{0.112\linewidth}
			\includegraphics[width=1.25\linewidth]{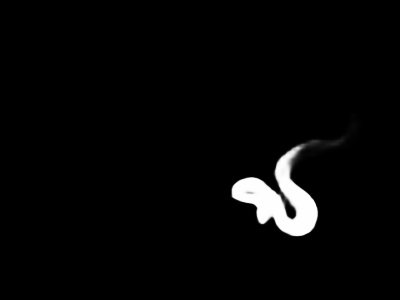}\vspace{1pt}	
			\includegraphics[width=1.25\linewidth]{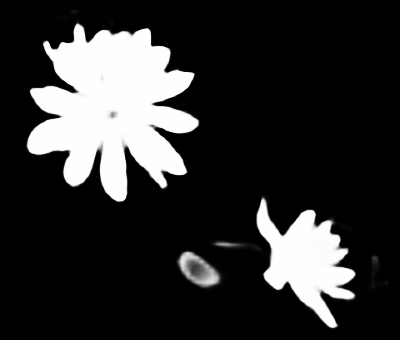}\vspace{1pt}
			\includegraphics[width=1.25\linewidth]{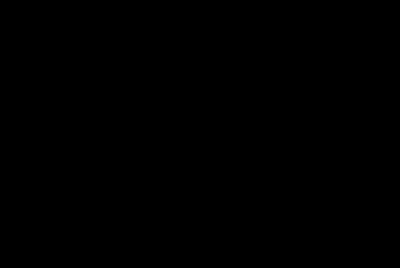}\vspace{1pt}			
			\includegraphics[width=1.25\linewidth]{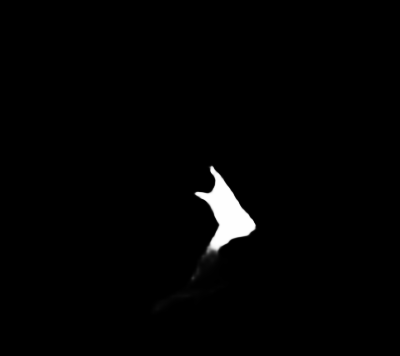}\vspace{1pt}	
	\end{minipage}}	
	\hspace{.07in}
	\subfigure[Baseline+H]{
		\begin{minipage}[b]{0.112\linewidth}			
			\includegraphics[width=1.25\linewidth]{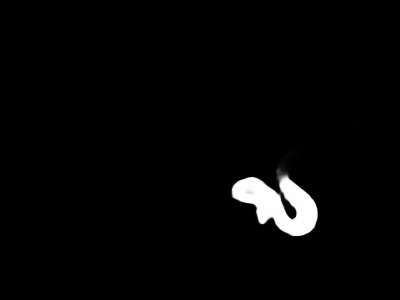}\vspace{1pt}	
			\includegraphics[width=1.25\linewidth]{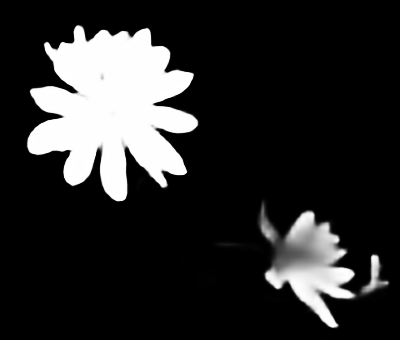}\vspace{1pt}
			\includegraphics[width=1.25\linewidth]{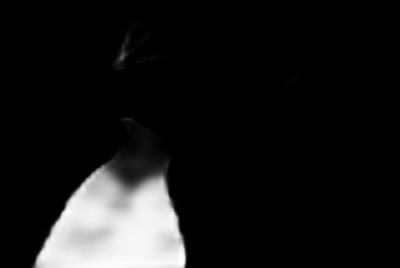}\vspace{1pt}			
			\includegraphics[width=1.25\linewidth]{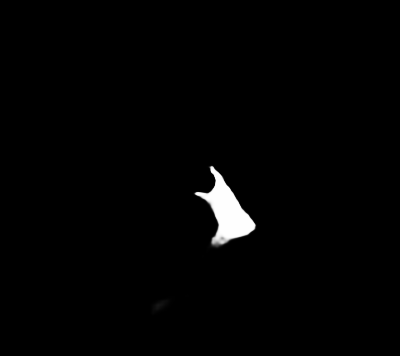}\vspace{1pt}	
	\end{minipage}}
	\hspace{.07in}
	\subfigure[Baseline+T]{
		\begin{minipage}[b]{0.112\linewidth}			
			\includegraphics[width=1.25\linewidth]{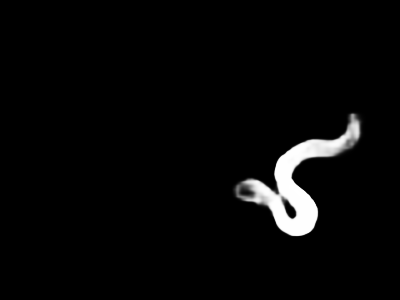}\vspace{1pt}		
			\includegraphics[width=1.25\linewidth]{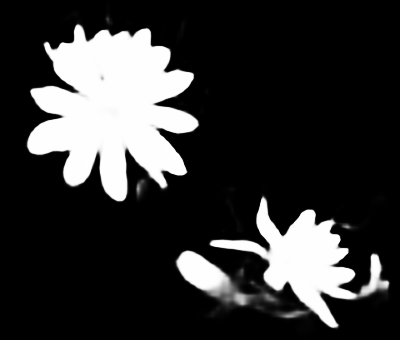}\vspace{1pt}
			\includegraphics[width=1.25\linewidth]{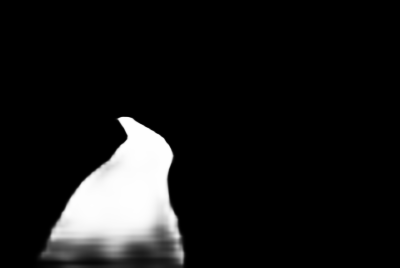}\vspace{1pt}			
			\includegraphics[width=1.25\linewidth]{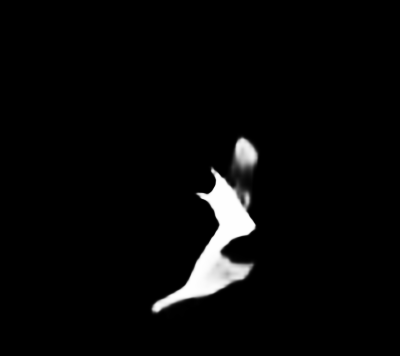}\vspace{1pt}			
	\end{minipage}}	
	\hspace{.07in}
	\subfigure[Baseline+T+H]{
		\begin{minipage}[b]{0.112\linewidth}			
			\includegraphics[width=1.25\linewidth]{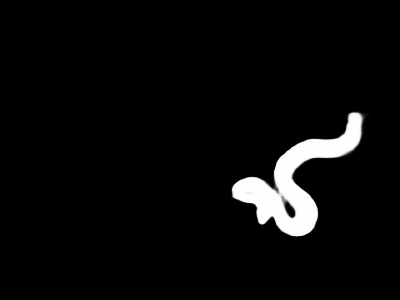}\vspace{1pt}	
			\includegraphics[width=1.25\linewidth]{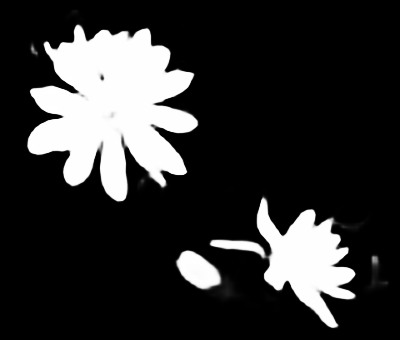}\vspace{1pt}
			\includegraphics[width=1.25\linewidth]{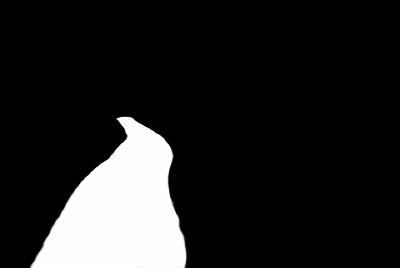}\vspace{1pt}				
			\includegraphics[width=1.25\linewidth]{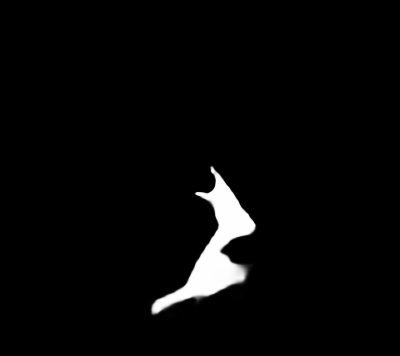}\vspace{1pt}			
	\end{minipage}}
	\hspace{.07in}
	\subfigure[T]{
		\begin{minipage}[b]{0.112\linewidth}			
			\includegraphics[width=1.25\linewidth]{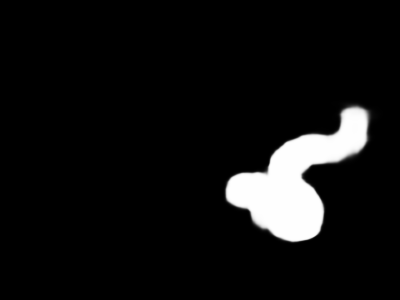}\vspace{1pt}	
			\includegraphics[width=1.25\linewidth]{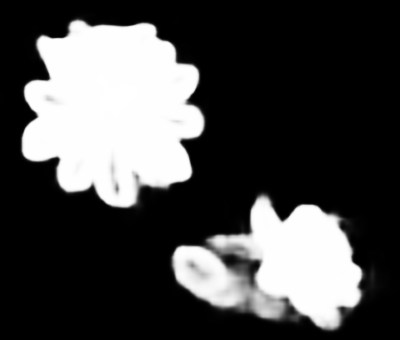}\vspace{1pt}
			\includegraphics[width=1.25\linewidth]{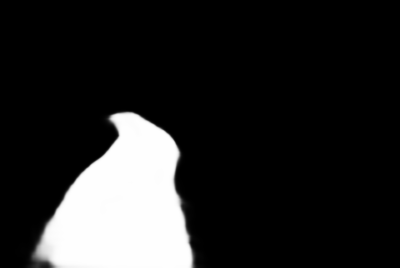}\vspace{1pt}				
			\includegraphics[width=1.25\linewidth]{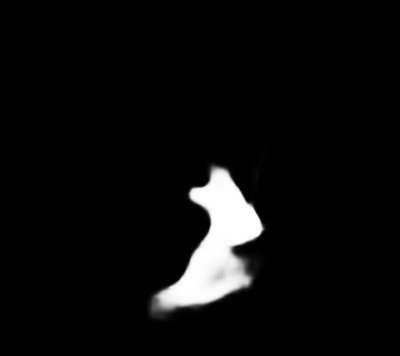}\vspace{1pt}				
	\end{minipage}}
	\caption{Visual comparisons for showing the benefits of the proposed components. Baseline+H: predictions of baseline with the proposed hierarchical difference aware loss. Baseline+T: predictions of baseline with TS branch. Baseline+T+H: predictions of our model which includes all proposed components. C: predictions of TS branch.}	
\end{figure*}
\subsection{Ablation Studies}
To investigate the effectiveness of different components of the proposed model, we conduct a series of experiments on the benchmark datasets to compare the performance variations of our model with different settings.
\subsubsection{Effectiveness of the Key Components}
To investigate the effectiveness of the proposed different components, we construct four models for comparisons. The first setting is baseline model where we remove the target separation (TS) branch and replace the proposed hierarchical difference aware (HDA) loss with combine of the BCE and IoU, denoted as  “Baseline”.  To validate the effectiveness of saliency features with expanded boundary (SF w/ EB) generated by TS branch and HDA loss, we conduct the second and third model by adding the TS branch (denoted as “Baseline + T”) and HDA (denoted as “Baseline + H”), respectively. The last setting is our model, which includes all proposed modules, denoted as “Baseline + T+ H”. The parameters of the four models are consistent. 
\par The comparison results of above mentioned models are listed in Tab. \uppercase\expandafter{\romannumeral3}. In particular, in terms of $F_{\beta}$, $S_m$ and MAE on the two massive datasets ,the “Baseline + H” gets additional $3.2\%$ improvement on average, and the “Baseline + T” gets additional $4.6\%$ improvement on average. Besides, we observe that the “Baseline + T” has better performance improvement than “Baseline + H”, which implies that the saliency features with expanded boundary is important for our model to obtain the complete prediction. As presented in Tab. \uppercase\expandafter{\romannumeral3}, our overall model (i.e., “Baseline + T + H”) achieves the better performance and gets additional $7\%$ improvement in terms of $F_{\beta}$, $S_m$ and MAE on average, which verifies the effectiveness of the proposed SF w/ EB and HDA loss.
\par In addition, as suggested by \cite{DBLP:conf/cvpr/WuFGWLD19}, we calculate the average score of the highlight pixel in saliency maps (ASHP). Given a predicted saliency map $P$, the ASHP could be given as:
\begin{equation}
{\mathop{\rm ASHP}\nolimits}  = \frac{1}{{{N_P}}}\sum\limits_{i = 1}^H {\sum\limits_{j = 1}^W {|P(i,j)} } |,
\end{equation}
where $N_{P}$ is the number of pixels which satisfies $P(i, j) \textgreater 0$. However, due to the lack of original ground truth as a measure, ASHP can only partially describe the highlight degree of model. Our experimental results show that when the predicted saliency map contains more noises or has a larger output size, the value of ASHP will also be higher. Therefore, we only use the ASHP / MAE (A/M) as an auxiliary metric in our model for ablation studies, which can evaluate the degree of noise and highlight of the model to a point. As presented in Tab. \uppercase\expandafter{\romannumeral3}, compared with “Baseline”, the “Baseline + T + H” gets additional $20.4\%$ improvement averagely in terms of A/M on the two massive datasets, which demonstrates our model can improve the highlight degree of predicted saliency maps under the premise of low background noises.
\begin{table}[ht]
	\centering
	\setlength{\abovecaptionskip}{0pt}
	\setlength{\belowcaptionskip}{3pt}
	\renewcommand\arraystretch{1.4}
	\setlength\tabcolsep{2pt}
	\caption{Ablation study for different modules on two massive benchmark datasets which have relatively cluttered background and high content variety. Baseline, T and H represent the baseline network, SF w/ EB generated by TS branch and HDA loss, respectively. The best results are highlighted in \textbf{bold}.}
	\begin{tabular}{l|cccc|cccc}
		\hline
		\multirow{2}{*}{Model} & \multicolumn{4}{c|}{DUT-OMRON}                & \multicolumn{4}{c}{DUTS-TE} 
		\\ \cline{2-9} 
		& $F_\beta$ & $S_m$ & MAE  & A/M  & $F_\beta$ & $S_m$ & MAE  & A/M 
		\\ \hline 
		Baseline            & 0.818     & 0.836  & 0.061  & 12.049 & 0.888   & 0.889 & 0.040 & 19.025
		\\ \hline
		Baseline + H            & 0.823     & 0.842  & 0.057  & 12.912 & 0.893   & 0.892 & 0.036 & 21.306
		\\ 
		Baseline + T            & 0.826     & 0.844  & 0.055 & 13.636  & 0.895   & 0.894 & 0.035 & 22.000
		\\ 
		\textbf{Baseline + T + H}            & \textbf{0.827}     & \textbf{0.847}  & \textbf{0.051}   & \textbf{14.569}   & \textbf{0.899}   & \textbf{0.896} & \textbf{0.034}     & \textbf{22.853}   
		\\ \hline
	\end{tabular}
\end{table}
\par Furthermore, because the improvement of  “Baseline + H” is smaller than “Baseline + T”, we train other model GCPA \cite{DBLP:conf/aaai/ChenXCH20} with different loss functions to further verify the effect of the proposed hierarchical difference aware loss. The comparison results are listed in Tab. \uppercase\expandafter{\romannumeral4}. As presented in Tab. \uppercase\expandafter{\romannumeral4}, compared with the other edge enhancement loss, the proposed hierarchical difference aware loss achieves the better performance. In particular, the performance is averagely improved by $7.1\%$ over the original BCE and IoU loss functions in terms of the $F_{\beta}$, $S_m$ and MAE, which further demonstrates the effectiveness of the proposed HDA loss. 
\begin{table}[ht]
	\centering
	\setlength{\abovecaptionskip}{0pt}
	\setlength{\belowcaptionskip}{3pt}
	\renewcommand\arraystretch{1.4}
	\setlength\tabcolsep{7pt}
	\caption{Comparison results with different loss functions trained on other model GCPA \cite{DBLP:conf/aaai/ChenXCH20}. The best results are highlighted in \textbf{bold}.}
	\begin{tabular}{l|ccc|ccc}
		\hline
		\multirow{2}{*}{Model} & \multicolumn{3}{c|}{DUT-OMRON}                & \multicolumn{3}{c}{DUTS-TE} 
		\\ \cline{2-7} 
		& $F_\beta$ & $S_m$ & MAE    & $F_\beta$ & $S_m$ & MAE   
		\\ \hline 
		BCE + IoU            & 0.815     & 0.834  & 0.057   & 0.876   & 0.880 & 0.041 
		\\ \hline
		PPA \cite{DBLP:conf/aaai/WeiWH20}          & 0.816     & 0.835  & 0.055   & 0.886   & 0.885 & 0.039 
		\\ \hline
		BRA \cite{zhu2021supplement}           & 0.819     & \textbf{0.841}  & 0.057   & 0.887   & 0.886 & 0.040 
		\\ \hline
		\textbf{HDA (Ours)}            & \textbf{0.821}     & \textbf{0.841}  & \textbf{0.052}   & \textbf{0.888}   & \textbf{0.888} & \textbf{0.037} 
		\\ \hline
	\end{tabular}
\end{table}
\par In addition, to further verify the SF w/ EB generated by expanded ground truth supervision, we replace the expanded ground truth with original ground truth in TS branch, denoted as "Baseline + O". This means the "Baseline + O" will output the standard saliency features. As presented in Tab. \uppercase\expandafter{\romannumeral5}, the performance of the "Baseline + T" is averagely improved by $2.4\%$ over the "Baseline + O" in terms of the $F_{\beta}$, $S_m$ and MAE, which further demonstrates the effectiveness of the proposed SF w/ EB. 
\begin{table}[ht]
	\centering
	\setlength{\abovecaptionskip}{0pt}
	\setlength{\belowcaptionskip}{3pt}
	\renewcommand\arraystretch{1.4}
	\setlength\tabcolsep{7pt}
	\caption{The effect of saliency features generated by different supervisory signals on performance. "Baseline + O" denotes that the original ground truth is used to supervise the TS branch, and "Baseline + T" denotes that the proposed expanded ground truth is used to supervise the TS branch. The best results are highlighted in \textbf{bold}.}
	\begin{tabular}{l|ccc|ccc}
		\hline
		\multirow{2}{*}{Model} & \multicolumn{3}{c|}{DUT-OMRON}                & \multicolumn{3}{c}{DUTS-TE} 
		\\ \cline{2-7} 
		& $F_\beta$ & $S_m$ & MAE    & $F_\beta$ & $S_m$ & MAE   
		\\ \hline
		Baseline + O            & 0.819     & 0.840 & 0.056   & 0.884   & 0.886 & 0.039 
		\\ \hline
		\textbf{Baseline + T}            & \textbf{0.826}     & \textbf{0.844}  & \textbf{0.055}   & \textbf{0.895}   & \textbf{0.894} & \textbf{0.035} 
		\\ \hline
	\end{tabular}
\end{table}
\subsubsection{Visual Comparisons of different Components}
The visual comparisons of prediction by different components are shown in Fig. 8. Compared with the “Baseline”, when the SF w/ EB generated by TS branch are introduced, the “Baseline + T”  is able to effectively improve the completeness of prediction. When the HDA loss is involved, “Baseline + T + H” can not only detect the complete salient objects, but also better suppress the noise. Besides, we show the expanded salient objects predicted by TS branch (column 7). Although the predictions of TS branch contain some noises, the structure of objects is relatively complete, which further verifies the effectiveness of the proposed model.

\section{Conclusion}
In this paper, we analysed that the methods based on supplementing information are difficult to detect complete salient objects from complex scenes, especially with cluttered background. To solve the issuses, we propose a novel network named SENet to generate complete saliency maps with low noise, which adopts separate expanded salient objects first, then segment fine objects. Furthermore, to identify the indistinguish regions between foreground and background, we propose an hierarchical difference aware loss that assigns weights to the pixels according to the distance from boundary in grades, which can further improve the prediction of structural integrity and local details. Compared with 15 state-of-the-art methods, including the method based on aggregating multi-level features and introducing edge or skeleton information, experimental results on five benchmark datasets show that the proposed model achieves superior performance over them. In addition, our model is more efficient with an inference speed 39.5 FPS on a single NVIDIA RTX 2080Ti GPU.


%



\section*{Acknowledgment}
This work was supported by the National Natural Science Foundation of China (NSFC) under Grant No. 62172243, National Key R\&D Program of China under Grant No. 2020YFB1710200, and Natural Science Foundation of Heilongjiang Province of China under Grant No. LH2022F044.

\ifCLASSOPTIONcaptionsoff
  \newpage
\fi



%
%
%
%

\bibliographystyle{IEEEtran}
\bibliography{Ourslib}
\end{document}